%% file: iclr2027_conference.tex
\documentclass{article} 

\usepackage[table,dvipsnames]{xcolor}

\usepackage{iclr2027_conference,times}

\input{math_commands.tex}

\usepackage{hyperref}
\usepackage{url}
\usepackage{graphicx}
\usepackage{booktabs}
\usepackage{multirow}

\usepackage{amsmath}
\usepackage{url}
\usepackage{tabularx} 
\usepackage{caption}
\usepackage{algorithm}
\usepackage{algorithmic}
\usepackage{float}
\usepackage{enumitem}
\usepackage[skins,breakable]{tcolorbox}
\usepackage{adjustbox}
\usepackage{siunitx}
\usepackage[table]{xcolor}
\usepackage{array}

\title{Just-In-Time Scene Graph Growth: Combating Perceptual Saturation in Long-Horizon Robotics}

\author{
\makebox[\linewidth][c]{%
{\normalfont Yue Chang\textsuperscript{1,*} \quad
Rufeng Chen\textsuperscript{1,*} \quad
Yifan Tian\textsuperscript{1,*} \quad
Dazhi Huang\textsuperscript{1} \quad
Zhaofan Zhang\textsuperscript{1}}%
} \\
\makebox[\linewidth][c]{%
{\normalfont Yi Chen\textsuperscript{2} \quad
Wenze Zhang\textsuperscript{1} \quad
Li Chen\textsuperscript{1} \quad
Hui Xiong\textsuperscript{1} \quad
Sihong Xie\textsuperscript{1,\textdagger}}%
} \\[2pt]
\makebox[\linewidth][c]{%
{\normalfont\small
\textsuperscript{1}The Hong Kong University of Science and Technology
(Guangzhou)}%
} \\
\makebox[\linewidth][c]{%
{\normalfont\small
\textsuperscript{2}Jilin University}%
} \\
\makebox[\linewidth][c]{%
{\normalfont\small
\textsuperscript{*}Equal contribution. \quad
\textsuperscript{\textdagger}Corresponding author.}%
}
}

\iclrfinalcopy 
\begin{document}

\maketitle
\fancyhead{}
\renewcommand{\headrulewidth}{0pt}

\input{Sections/0_abs}
\input{Sections/1_Intro}
\input{Sections/2_related_new}
\input{Sections/3_motivation_exp}
\input{Sections/4_method_new}
\input{Sections/5_data_bench}
\input{Sections/6_exp}
\input{Sections/7_conclusion}

\subsection*{AI Use Statement}

We used generative AI tools in three supporting roles during the preparation
of this work.
First, AI-assisted writing tools were used to improve grammar, clarity, and
presentation of author-written text.
Second, generative AI was used to propose candidate task annotations and
instructions for JITOMA-Bench; all generated candidates were subsequently
reviewed, corrected when necessary, and approved by human annotators according
to the annotation protocol described in the paper.
Third, generative AI was used to generate preliminary visual concepts and
layout ideas that assisted the authors in designing several figures; the final
figures and scientific content were produced and verified by the authors.
Generative AI was not used to fabricate experimental measurements or results,
replace human validation of benchmark annotations, or make final scientific
judgments regarding the methods, analyses, or conclusions of this work.
All AI-assisted text, annotations, and visual materials were manually reviewed
for correctness, consistency, and appropriate attribution.
The authors take full responsibility for the final content of this paper,
including all text, claims, data, experiments, and artifacts produced with the
aid of generative AI.




\subsection*{Reproducibility Statement}

We provide sufficient details to reproduce JITOMA and JITOMA-Bench.
The JITOMA pipeline is specified in Sec.~\ref{sec:method}, while benchmark
construction, task annotations, and evaluation metrics are described in
Sec.~\ref{sec:benchmark}, with dataset statistics and annotation details in
Appendix~\ref{app:jitoma_bench_details}.
The controlled perceptual-saturation study, including prompt construction,
distractor controls, and evaluation protocol, is given in
Appendix~\ref{supp:perceptual_saturation_controlled}.
Implementation settings, memory-management rules, default hyperparameters,
and sensitivity analysis are provided in
Appendix~\ref{app:deployment_details}, and experimental baselines and model
configurations in Sec.~\ref{sec:experiments}.
Upon acceptance, we will release the implementation, benchmark annotations,
and evaluation scripts to facilitate reproducibility and future research.



\bibliography{iclr2027_conference}
\bibliographystyle{iclr2027_conference}

\clearpage
\appendix
\section{Appendix}
\input{Sections/Appendix/3_bench}
\input{Sections/Appendix/0_moti_exp}
\input{Sections/Appendix/1_hyper}

\end{document}

%% file: math_commands.tex
\usepackage{amsmath,amsfonts,bm}

\def\eqref#1{equation~\ref{#1}}

\def\1{\bm{1}}

\DeclareMathAlphabet{\mathsfit}{\encodingdefault}{\sfdefault}{m}{sl}
\SetMathAlphabet{\mathsfit}{bold}{\encodingdefault}{\sfdefault}{bx}{n}



%% file: Sections/0_abs.tex
\begin{abstract}
While 3D Scene Graphs (3DSGs) provide crucial structured representations for embodied agents, conventional Ahead-of-Time, ``build-everything-then-filter'' pipelines conflict with the real-time, low-latency demands of edge platforms, inducing a \textit{perceptual saturation effect} via severe observation redundancy. 
To resolve this, we present \textbf{JITOMA} (\textbf{J}ust-\textbf{I}n-\textbf{T}ime \textbf{O}n-demand \textbf{M}emory \textbf{A}ctivation), a closed-loop framework that unifies task reasoning, perception, and memory into a just-in-time growth process. 
Instead of exhaustively mapping the entire environment, JITOMA leverages a top-down task heatmap at the frontend to filter continuous observations, routing minimal streams to maintain a global foundation of low-cost, \textit{dormant anchors}. 
Upon a cognitive query, the backend Large Language Model (LLM) parses the robotic intent to dynamically awaken task-relevant anchors, triggering expensive semantic operations such as dense node captioning
exclusively within the activated local subgraph.
To evaluate these dynamic capabilities and study perceptual saturation trade-offs, we introduce \textbf{JITOMA-Bench}, a benchmark for long-horizon task switching and complex intent grounding.
Across JITOMA-Bench, JITOMA maintains only $1$--$6$ active semantic objects
and $0.25$--$0.28$ s/frame across all tiers, showing that semantic computation
remains bounded by current task demand rather than accumulated scene complexity.
\end{abstract}

%% file: Sections/1_Intro.tex
\section{Introduction}
The ability to perceive, memorize, and reason about 3D environments is
central to embodied intelligence.
3D Scene Graphs (3DSGs) provide a structured representation for this purpose,
encoding objects as nodes and their relationships as edges.
With the advent of Vision-Language Models (VLMs), 3DSGs have evolved from
closed-set semantics
\citep{hughes2022hydra, rosinol2021kimera, wu2021scenegraphfusion}
to open-vocabulary representations
\citep{fungraph3d, gu2024conceptgraphs, jatavallabhula2023conceptfusion,
koch2024open3dsg, linok2025beyond, maggio2024clio, werby2024hierarchical,
yamazaki2024open, yan2025dynamic, chang2026rag},
enabling zero-shot querying of novel objects.
More recently, task-driven
\citep{maggio2024clio, maggio2026found, agia2022taskography,
maggio2025bayesian}
and functional graphs
\citep{fungraph3d, ju2025momagraph, buchner2026articulated,
rotondi2025fungraph}
have begun to reshape graph construction around downstream robotic tasks.
These advances support increasingly rich spatial memory for robot navigation
\citep{werby2024hierarchical, yan2025dynamic, gadre2022clip, shah2023lm,
yin2024sg, chen2026psg, xia2026exploring}
and manipulation
\citep{yan2025dynamic, shridhar2022cliport, rashid2023language,
honerkamp2024language}.

However, most existing 3DSG pipelines still follow an Ahead-of-Time (AOT),
``build-everything-then-filter'' construction process, instantiating semantic
structure before knowing which parts will matter to the downstream task.
As shown by our diagnostic study in Sec.~\ref{sec:preliminary_findings},
this indiscriminate accumulation introduces
\textbf{observation redundancy} and can lead to a
\textit{\textbf{perceptual saturation effect}}, where additional global
semantics no longer improve downstream reasoning and may instead increase
semantic clutter and processing cost.
Importantly, recent task-driven methods
\citep{maggio2024clio, ju2025momagraph}
condition graph structure or granularity on downstream tasks, but operate
over task-agnostic primitives or acquired visual observations.
Thus, task intent primarily shapes \emph{what} is represented, rather than
\emph{where and when} computation is allocated across perception and memory.

Humans offer a useful analogy: rather than resolving every observed object in
full detail, perception retains lightweight impressions and allocates
fine-grained processing only when a concrete intent demands it.
Inspired by this resource-aware behavior, we introduce
\textbf{just-in-time (JIT) scene graph growth}.
Analogous to JIT compilation, semantic structure is instantiated only at
runtime where and when it is needed, rather than accumulated globally in
advance.
This reframes scene graph construction as a process of selective resource
allocation: task-irrelevant observations remain lightweight and dormant,
while computation is concentrated on task-relevant entities on demand.

Following this principle, we introduce \textbf{JITOMA}
(\textbf{J}ust-\textbf{I}n-\textbf{T}ime \textbf{O}n-demand
\textbf{M}emory \textbf{A}ctivation), an online, closed-loop 3DSG framework
that couples task intent with perception and memory.
At the frontend, top-down task heatmaps gate continuous observations before
memory commitment, while stable observations are retained as lightweight
\emph{dormant anchors} without explicit semantic expansion.
When a command arrives, JITOMA retrieves and activates only task-relevant
anchors, invoking expensive operations such as dense captioning on demand and,
when needed, further functional refinement.
After execution, reusable semantics are distilled back into dormant memory and
the active state collapses.
As tasks change, computation therefore shifts between small task-relevant
subsets of a persistent memory foundation, rather than accumulating globally
over time.

To evaluate these process-level capabilities, we introduce
\textbf{JITOMA-Bench} (Sec.~\ref{sec:benchmark}), a three-tier benchmark
extending Clio-Bench \citep{maggio2024clio} with continuous real-world trajectories and covering
foundational grounding, long-horizon task switching, and complex intent
grounding.
JITOMA-Bench jointly evaluates grounding accuracy and online efficiency.
Across all tiers, JITOMA maintains only $1$--$6$ active semantic objects and
$0.25$--$0.28$\,s/frame while preserving strong grounding accuracy.
This demonstrates that semantic computation can remain bounded by current
task demand without sacrificing task-relevant grounding under continual task
switching.

%% file: Sections/2_related_new.tex
\section{Related Work}
\label{sec:related_work}

\textbf{Ahead-of-Time Scene Graph Construction.}
3D Scene Graphs (3DSGs) organize sensory observations into structured object
nodes and semantic relationships for embodied spatial reasoning
\citep{hughes2022hydra, rosinol2021kimera}.
Early systems largely relied on closed-set perception
\citep{hughes2022hydra, rosinol2021kimera, wu2021scenegraphfusion},
while recent methods increasingly incorporate open-vocabulary semantics through
vision-language models
\citep{fungraph3d, gu2024conceptgraphs, jatavallabhula2023conceptfusion,
koch2024open3dsg, linok2025beyond, maggio2024clio, werby2024hierarchical,
yamazaki2024open, yan2025dynamic, chang2026rag}.
Contemporary approaches further enrich scene graphs with hierarchical
organization and functional knowledge to support downstream interaction
\citep{fungraph3d, ju2025momagraph, buchner2026articulated}.
Despite these advances in representation, most pipelines still follow an
Ahead-of-Time (AOT) construction pattern: visual evidence and semantic
structure are progressively instantiated during mapping before their relevance
to a future task is known.
This design provides rich persistent memory, but allocates perception and
semantic computation broadly across the scene.
As exploration grows, task-irrelevant observations and semantics therefore
accumulate together with useful content, creating the observation redundancy
and perceptual saturation studied in this work.

\textbf{Task-Driven Representations and Just-In-Time Growth.}
A growing line of work instead conditions scene representations on downstream
tasks.
Taskography sparsifies existing 3DSGs for tractable task planning
\citep{agia2022taskography};
Clio clusters task-agnostic 3D primitives into task-relevant objects and
regions \citep{maggio2024clio};
Bayesian Fields introduces task-dependent granularity into open-set semantic
mapping \citep{maggio2025bayesian};
and MomaGraph constructs task-oriented graph structure from acquired visual
observations \citep{ju2025momagraph}.
These methods demonstrate the benefit of adapting graph content or granularity
to downstream requirements, but task conditioning generally operates after
visual evidence or low-level primitives have already been acquired.
FOUND-IT \citep{maggio2026found} moves further toward on-demand construction
by retaining a visual memory and resolving object and region granularity when
runtime tasks arrive.
JITOMA builds on this direction from a complementary perspective: task intent
is used not only to determine \emph{what} should be represented, but also to
govern \emph{where and when} computation is allocated across perception and
memory throughout online scene construction.

%% file: Sections/3_motivation_exp.tex
\section{Motivation: Perceptual Saturation in AOT Scene Graphs}
\label{sec:preliminary_findings}

Conventional Ahead-of-Time (AOT) scene graph construction pursues increasingly
complete global representations.
We ask whether this exhaustive accumulation is always beneficial for
reasoning, or whether excessive task-irrelevant semantics can instead become a
source of cognitive burden.

\paragraph{Does global structural completeness overwhelm downstream reasoners?}
We conduct a diagnostic study following the complex-intent setting of
Tier~3 in JITOMA-Bench (Sec.~\ref{sec:benchmark}).
A Large Language Model (Qwen3.5-9b~\citep{qwen3.5}) plans the command
``pack my eyeglasses for travel'' in a cluttered real-world cubicle.
We compare two upstream representations:
\textbf{AOT Full Graph}, containing all $18$ environmental entities constructed
by a task-agnostic pipeline, and
\textbf{JIT Subgraph (Ours)}, containing only the $3$ task-relevant entities
(\emph{eyeglasses}, \emph{glasses case}, and \emph{backpacks})
activated on demand.
Both conditions use the identical planner and prompt, differing only in the
scene-graph entities exposed to the planner
(Appendix~\ref{supp:prompt_aot_motivation}).

\begin{figure}[ht]
  \centering
  \includegraphics[width=\linewidth]{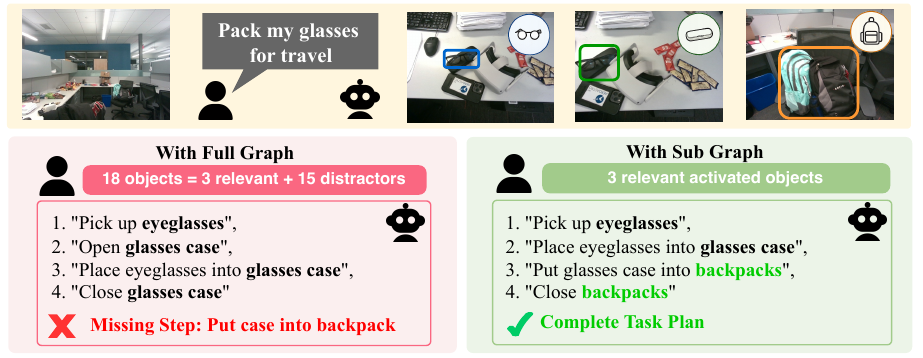}
  \caption{
\textbf{Illustrative example of perceptual saturation.}
\textbf{Left:} AOT exposes all $18$ entities, including $15$ distractors,
causing the planner to miss the final containment step.
\textbf{Right:} JIT exposes only $3$ task-relevant entities, enabling the same planner to recover the complete execution sequence.
}
  \label{fig:perceptual_saturation}
\end{figure}

As shown in Fig.~\ref{fig:perceptual_saturation}, the AOT graph exposes the
planner to $15$ task-irrelevant distractors (e.g., \emph{hardware drills}
and \emph{mudstone rocks}).
The planner terminates after placing the eyeglasses into the case and omits
the final step of placing the case into the backpack, whereas the same planner
produces the complete sequence when given only the three task-relevant
entities.

To examine whether this failure mode extends beyond a single example, we
further conduct a controlled distractor study over all $25$ Tier~3 tasks
(Appendix~\ref{supp:perceptual_saturation_controlled}).
Holding the task-relevant graph content and planner prompt fixed, we
progressively add $k\in\{0,5,10,15\}$ task-irrelevant entities.
Across 1,194 planner requests, mean all-target coverage decreases from
82.31\% to 75.24\%, while latent-target coverage drops more substantially
from 68.10\% to 56.30\%.
In contrast, primary-target coverage remains nearly unchanged
(98.52\% vs.\ 98.11\%), suggesting that irrelevant semantic load primarily
disrupts the recovery of implicit objects required for task execution.

This observation motivates controlling semantic load before it reaches the
downstream reasoner. While task-driven filtering can mitigate clutter at
query time, JITOMA moves this selectivity upstream: task intent governs where
perceptual and semantic computation are allocated during scene construction.
Taken together, the qualitative example and controlled study support a
\textbf{\textit{perceptual saturation effect}}: when task-relevant information
is held fixed, increasing irrelevant semantic load can impair the recovery of
task-required information, particularly latent dependencies.


\begin{tcolorbox}[
    colback=orange!5!white,
    colframe=orange!60!black,
    title=Finding: Perceptual Saturation Effect,
    fonttitle=\bfseries,
    arc=1mm,
    breakable
]
\footnotesize
With task-relevant graph content held fixed, increasing task-irrelevant
semantic load can reduce downstream target coverage, with the strongest
degradation on latent objects required for task completion.
\end{tcolorbox}

%% file: Sections/4_method_new.tex
\section{Method}
\label{sec:method}

\subsection{Problem Formulation and Architecture Overview}
\label{sec:method_overview}

\begin{figure*}[ht]
    \centering
    \includegraphics[width=\textwidth]{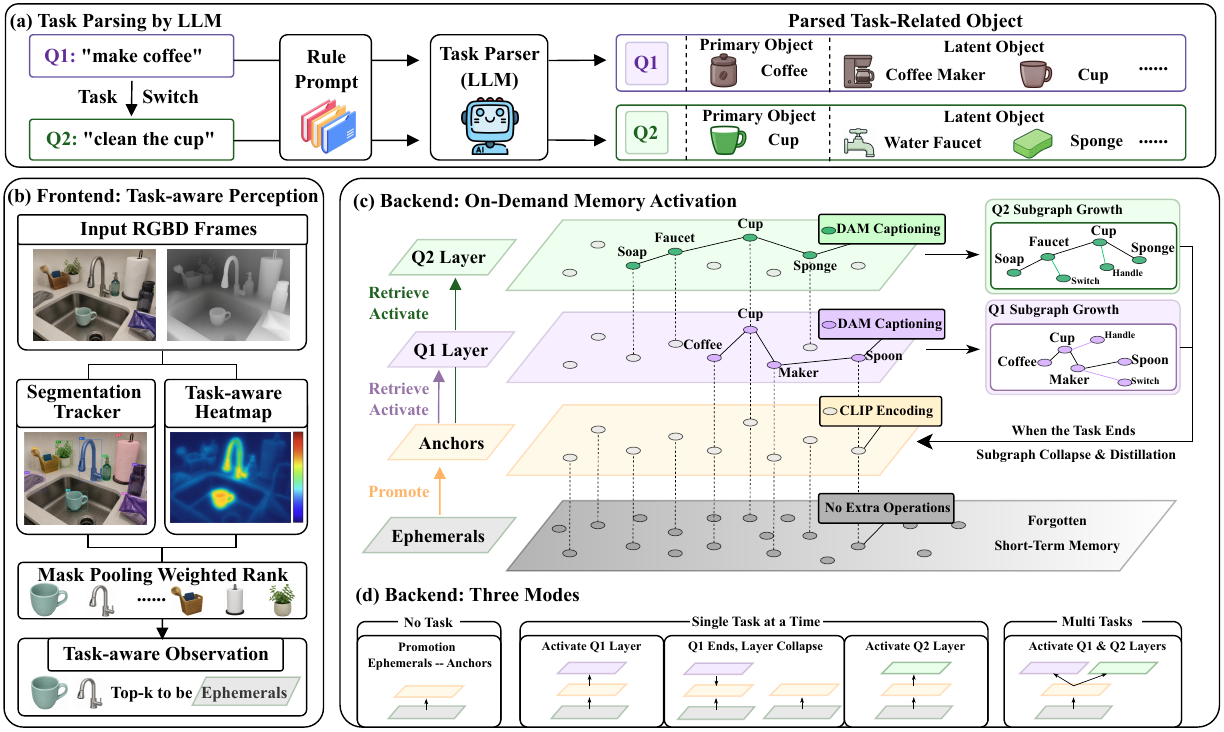}
    \caption{\textbf{Overview of JITOMA.}
\textbf{(a) Intent Parsing:} An LLM extracts primary and latent targets.
\textbf{(b) Frontend Perception:} Task heatmaps gate RGB-D observations before
memory commitment.
\textbf{(c) Backend JIT Activation:} Dormant anchors are selectively activated
for semantic expansion and optional functional refinement.
\textbf{(d) Operating Modes:} JITOMA supports task-free writing, single-query
activation, and concurrent queries over a shared dormant memory.}
    \label{fig:pipeline}
\end{figure*}

JITOMA targets online 3D scene graph construction under long-horizon embodied execution, where conventional Ahead-of-Time (AOT) pipelines instantiate global semantic structures before knowing which parts of the scene will be used.
Given a continuous RGB-D stream and camera poses $\{(I_t,D_t,P_t)\}_{t=1}^{T}$, JITOMA instead treats graph construction as a \emph{just-in-time growth process}: exploration maintains a sparse memory foundation, while expensive semantic computation is triggered only when required by the current task.
At time $t$, JITOMA maintains a two-tier memory state
\begin{equation}
    \mathcal{M}_t
    =
    \big(V_t^{\mathrm{eph}},V_t^{\mathrm{anc}}\big),
    \label{eq:memory_state}
\end{equation}
where $V_t^{\mathrm{eph}}$ contains ephemeral short-term hypotheses and
$V_t^{\mathrm{anc}}$ contains long-lived dormant anchors.
Ephemeral hypotheses represent transient observations, whereas dormant
anchors retain stable observational evidence without explicit semantics such
as object captions.
Upon a command $q$, JITOMA activates only task-relevant anchors, materializes their semantic
structure on demand, and distills reusable semantics back into dormant memory
after execution.
Thus, expensive graph structure follows active cognitive demand rather than
accumulating with exploration time.

\subsection{Task-Guided Perception and Memory Formation}
\label{sec:method_frontend}

Unlike task-driven representations operating on acquired evidence, JITOMA
lets task intent govern memory formation before retrieval.
Given command $q$, an LLM parser extracts
$C(q)=C_{\mathrm{pri}}(q)\cup C_{\mathrm{lat}}(q)$,
where $C_{\mathrm{pri}}$ denotes explicit targets and $C_{\mathrm{lat}}$
execution-implied objects.
For example, Fig.~\ref{fig:pipeline}(a) parses ``clean the cup'' into
$C_{\mathrm{pri}}=\{\emph{cup}\}$ and
$C_{\mathrm{lat}}=\{\emph{water faucet},\emph{sponge}\}$.

At timestamp $t$, a class-agnostic segmentation tracker~\citep{gorlo2026describe}
produces $O_t=\{o_{t,i}\}_{i=1}^{N_t}$, with
$o_{t,i}=(M_{t,i},x_{t,i},g_{t,i})$ denoting its 2D mask, crop, and coarse
3D geometry.
Before semantic labeling, the parsed concepts produce a top-down task heatmap
$H_t^q$ through a vision-language alignment model~\citep{clipseg}, and only
the most relevant observations are admitted:
\begin{equation}
    \widetilde{O}_t(q)
    =
    \operatorname{TopK}_{o_{t,i}\in O_t}
    \operatorname{Pool}\!\left(H_t^q\odot M_{t,i}\right).
    \label{eq:frontend_gating}
\end{equation}
This gate acts before memory commitment, suppressing redundant observations
rather than filtering a fully constructed graph.
When no task is active, JITOMA uses conservative writing, admitting new tracks
and retaining repeated observations only when visual evidence improves
(Appendix~\ref{app:conservative_writing}).

\paragraph{From Ephemeral Hypotheses to Dormant Anchors.}
Admitted observations $\widetilde{O}_t(q)$ enter the \emph{Ephemerals} layer
as short-term hypotheses and accumulate lightweight visual and geometric
evidence.
A hypothesis is promoted to a dormant anchor $v_j\in V_t^{\mathrm{anc}}$
after sufficient observations; otherwise, it expires.
At promotion, JITOMA stores a representative crop $\bar{x}_j$ and visual key
$z_j=\operatorname{CLIP}_{\mathrm{img}}(\bar{x}_j)$.
Anchors retain stable evidence but remain semantically dormant, so persistent
memory grows without globally materializing object semantics.

\subsection{On-Demand Memory Activation}
\label{sec:method_activation}

When a command $q$ arrives, JITOMA activates memory through a lightweight retrieval stage followed by selective semantic verification.
For readability, we omit the time subscript and operate on the current dormant anchor set $V^{\mathrm{anc}}$.
For each concept $c\in C(q)$, candidate anchors are retrieved using the CLIP visual key together with accumulated frontend heatmap evidence:
\begin{equation}
    B(c)
    =
    \operatorname{TopK}_{v_j\in V^{\mathrm{anc}}}^{K_{\mathrm{act}}}
    \left[
        \cos\!\left(\operatorname{CLIP}_{\mathrm{text}}(c),z_j\right)
        +\lambda\tilde{h}_j^c
    \right].
    \label{eq:candidate_retrieval}
\end{equation}
Here, $\tilde{h}_j^c$ denotes the accumulated task-heatmap evidence associated with anchor $v_j$.
Only anchors in $B(c)$ invoke expensive semantic processing; all others remain dormant.
JITOMA batch-captions the retrieved candidates with DAM~\citep{gorlo2026describe,dam}, producing descriptions $d_j$, and then uses an LLM reranker to select the best anchor for each concept:
\begin{equation}
    V^{\mathrm{act}}(q)
    =
    \left\{
        \operatorname{LLM}_{\mathrm{rerank}}
        \!\left(q,c,\{(v_j,d_j)\}_{v_j\in B(c)}\right)
        \;\middle|\;
        c\in C(q)
    \right\}
    \subseteq V^{\mathrm{anc}}.
    \label{eq:activation}
\end{equation}
Unselected candidates return immediately to the dormant state.
Thus, captioning and language reasoning are restricted to a small task-relevant candidate set instead of being applied globally.

\subsection{Just-In-Time Semantic Growth}
\label{sec:method_growth}

Semantic growth occurs only after activation: selected anchors
$V^{\mathrm{act}}(q)$ materialize explicit object semantics $d_j$, while the
rest of memory remains dormant. Thus, JITOMA expands only task-relevant
objects, confining semantic computation to current task demand.

\paragraph{Optional functional refinement.}
Beyond the object-level semantic activation evaluated in this work, JITOMA
can optionally refine an activated object with a task-conditioned
interaction part when finer-grained physical interaction is required.
Given an activated anchor $v_j$, its representative crop $\bar{x}_j$,
caption $d_j$, and command $q$, a VLM predicts the interaction-relevant part,
which is grounded in 2D and back-projected into 3D:
\begin{equation}
    \ell_j^q
    =
    \operatorname{VLM}_{\mathrm{part}}(q,\bar{x}_j,d_j),
    \qquad
    b_j^q
    =
    \operatorname{OBB}\!\left(
        \Pi^{-1}\!\left(
            \operatorname{Seg}(\bar{x}_j,\ell_j^q),
            D_j,P_j
        \right)
    \right).
    \label{eq:functional_grounding}
\end{equation}
Here, $\ell_j^q$ denotes the interaction-relevant part and $b_j^q$ its
localized 3D extent; the resulting node $p_j^q=(b_j^q,\ell_j^q)$ is attached
only to the activated anchor.
This optional extension applies the JIT principle to finer-grained
structure, while our evaluation focuses on semantic
activation and grounding.

\subsection{Semantic Distillation and Operating Modes}
\label{sec:method_distillation}

After task execution, JITOMA collapses the active semantic state and distills
reusable outputs into the corresponding dormant anchors.
For each $v_j\in V^{\mathrm{act}}(q)$, its caption $d_j$ and, when available,
functional refinement $p_j^q$ are retained as auxiliary attributes via
$\mathcal{A}_j\leftarrow\mathcal{A}_j\cup\{d_j,p_j^q\}$.
Temporary heatmaps, masks, retrieval candidates, and task-specific overlays are
discarded, while distilled semantics remain dormant until reactivated by a
subsequent query.

As illustrated in Fig.~\ref{fig:pipeline}(d), JITOMA supports three modes.
With no query, it performs conservative writing and anchor promotion.
With a single query, relevant anchors are activated, semantically expanded,
distilled after execution, and returned to the dormant state.
With concurrent queries, multiple task-specific overlays share the same
dormant anchor foundation.

%% file: Sections/5_data_bench.tex
\section{Dataset and Benchmark}
\label{sec:benchmark}

Clio-Bench~\citep{maggio2024clio} provides real-world RGB-D trajectories
and geometric ground truth for evaluating task-driven open-set 3DSGs.
However, existing benchmarks overlook three aspects:
(G1) \textbf{Peak Active-State Dynamics}, as final graph statistics do not
capture online peak semantic-state growth and computation;
(G2) \textbf{Long-Horizon Multi-Tasking}, as evaluation rarely considers
continuous task switching within the same scene;
and (G3) \textbf{Complex Intent Grounding}, as commands involve only
explicit target objects rather than multiple primary and latent entities.

To address these gaps, we introduce \textbf{JITOMA-Bench}, extending Clio's
three scenes with a three-tier evaluation protocol covering single-query
grounding, multi-task adaptation, and complex intent
grounding (Sec.~\ref{sec:benchmark_settings}).
JITOMA-Bench contains 126 evaluation tasks,
156 task-target references, and 80 unique annotated 3D boxes, with detailed
statistics provided in Appendix~\ref{app:jitoma_bench_details}.

\begin{figure*}[ht]
    \centering
    \includegraphics[width=\textwidth]{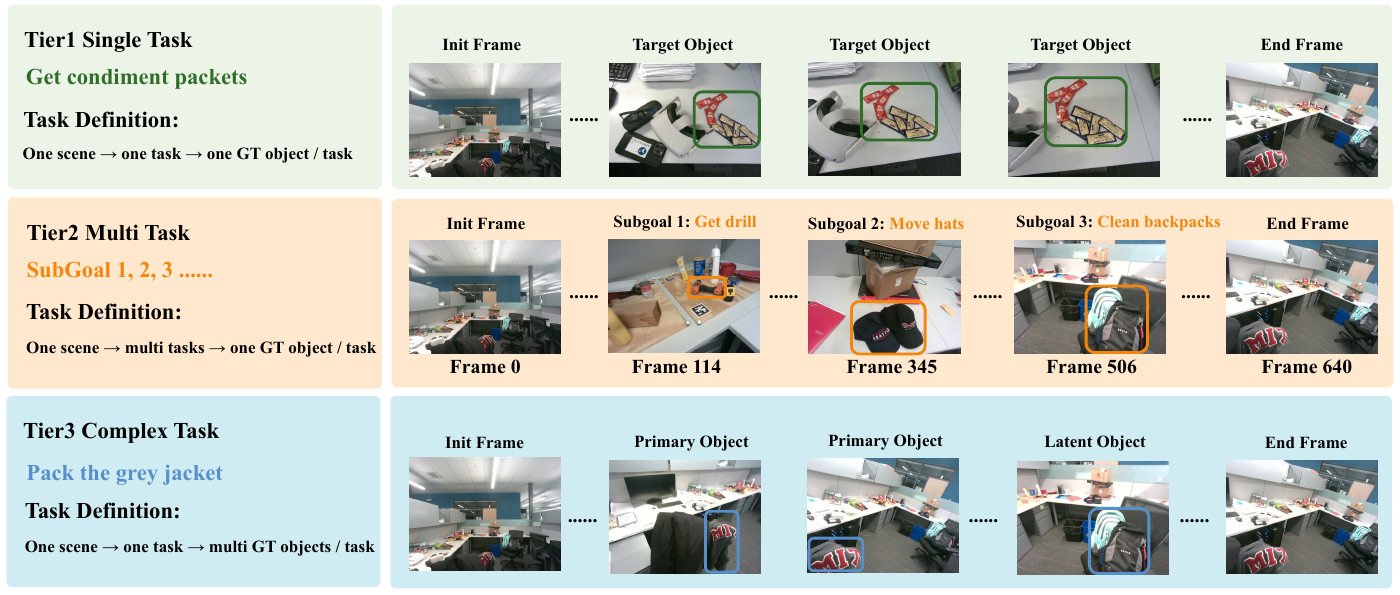}
    \caption{\textbf{Overview of the three-tier JITOMA-Bench evaluation.}
    The benchmark progressively evaluates explicit single-object grounding (Tier~1), long-horizon task switching within a continuous scene (Tier~2), and complex instruction grounding over multiple primary and latent objects (Tier~3).}
    \label{fig:bench}
\end{figure*}

\subsection{Task Annotations and Complementary Evaluation Tracks}
\label{sec:benchmark_settings}


\noindent\textbf{Tier 1: Foundational Grounding.}
As shown in Fig.~\ref{fig:bench}, Tier~1 evaluates each explicit single-GT task independently, providing a controlled measure of conventional 3DSG object grounding.

\noindent\textbf{Tier 2: Long-Horizon Dynamics.}
Tier~2 addresses G2 by evaluating continual adaptation under an evolving task stream within the same scene.
To construct potentially overlapping tasks, we manually annotate the first- and last-visible frames of each target (e.g., Frame X for Subgoal x in Fig.~\ref{fig:bench}).
Each task is activated at its first-visible frame and evaluated ten frames after its last-visible frame.

\noindent\textbf{Tier 3: Complex Intent Grounding.}
Tier~3 evaluates instructions involving multiple explicit and latent objects
following the Tier~1 timing protocol.
Using Clio ground-truth labels, GPT-5.4~\citep{openai2026gpt54} generates
candidate tasks for the Apartment, Office, and Cubicle scenes by composing
single-object tasks into multi-object instructions.
The authors assign primary and latent objects according to Clio labels, and
three PhD researchers independently review each instruction.
Candidates are removed if at least two reviewers identify ambiguity or unnecessary objects for task completion.

\subsection{Evaluation Metrics}
\label{sec:benchmark_metrics}

Our evaluation follows two complementary objectives: \textbf{grounding accuracy} and \textbf{system efficiency}. 
For accuracy, we report the top-1 Intersection over Union (IoU), averaged across queries, to
measure geometric agreement between the retrieved object and its ground truth. We further define
mR@K as the arithmetic mean of Recall@K over IoU thresholds
$\{0.1, 0.2, 0.3\}$, capturing both semantic retrieval and localization robustness.
In Tier~3, Primary and Latent concepts are evaluated separately before being averaged within each task, ensuring that performance reflects both explicit-object grounding and implicit-intent understanding.

For efficiency, we distinguish the \emph{active semantic state} from the
persistent visual memory.
\textbf{Objs} counts the semantic object nodes active at query time, while
\textbf{Peak} is the maximum active semantic-node count over the scene replay.
For JITOMA, dormant anchors are excluded from these active-state counts and
reported separately in parentheses; methods without an explicit
dormant/active split count all nodes in their native semantic representation.
Thus, Objs and Peak characterize semantic working-set complexity rather than
byte-level memory footprint.
Finally, \textbf{TPF} (Time Per Frame) measures the average processing time
per frame.


%% file: Sections/6_exp.tex
\section{Experiments}
\label{sec:experiments}

We evaluate JITOMA on JITOMA-Bench (Sec.~\ref{sec:benchmark}), covering
single-query grounding, long-horizon task switching, and complex
Primary--Latent intent grounding.
We compare JITOMA against open-vocabulary and task-driven 3DSG
baselines under both grounding accuracy and efficiency metrics.

\subsection{Experimental Setup}
\label{sec:exp_setup}

\noindent\textbf{Baselines.}
We benchmark JITOMA against three representative 3DSG frameworks:
\textbf{ConceptGraphs}~\citep{gu2024conceptgraphs}, an open-vocabulary
scene graph method that constructs a global semantic representation by
aggregating object-level features;
\textbf{ReasoningGraph}~\citep{puigjaner2026relationship}, a hierarchical
scene graph framework that augments object representations with relational
information for structured scene understanding;
and \textbf{Clio}~\citep{maggio2024clio}, a real-time task-driven scene graph
method that derives task-relevant representations from open-world 3D
primitives.

\noindent\textbf{Implementation Details.}
We implement JITOMA using Qwen3.5-9b~\citep{qwen3.5} as the foundation model,
paired with DAM~\citep{dam} in batch mode~\citep{gorlo2026describe} for
on-demand captioning.
All implementation details and configurations are provided in
Appendix~\ref{app:deployment_details}.

\input{Tables/Table1_main_result}

\subsection{Main Results: Accuracy and System Efficiency}
\label{sec:exp_main_results}

We report the overall evaluation on JITOMA-Bench in
Table~\ref{tab:main_results}.
JITOMA achieves strong grounding accuracy while maintaining a compact
active representation and stable online processing cost across increasingly
challenging task settings.

\noindent\textbf{Grounding accuracy under diverse task settings.}
JITOMA achieves strong grounding performance across all three evaluation
tiers, demonstrating robustness from single-object retrieval to complex
Primary--Latent intent grounding.
On Tier~1, JITOMA obtains the best overall accuracy in most scenes, e.g.,
20.4 IoU / 53.0 mR@1 / 60.6 mR@3 in Office and
21.9 IoU / 55.6 mR@1 / 66.7 mR@3 in Cubicle.
On Tier~2, it maintains high retrieval accuracy under long-horizon task
streams, achieving 29.1 IoU / 64.3 mR@1 / 73.8 mR@3 in Office.
For Tier~3, where instructions require both explicit and latent object
grounding, JITOMA achieves the highest accuracy across all scenes, reaching
19.8 IoU / 50.7 mR@1 / 58.0 mR@3 in Office.
These results suggest that task-conditioned memory formation helps reduce
irrelevant observations before semantic expansion, enabling retrieval from a
more compact task-relevant representation.

\noindent\textbf{Bounded semantic growth and online efficiency.}
Table~\ref{tab:main_results} separates JITOMA's active semantic state from its
dormant visual memory: Objs/Peak count active semantic nodes, while
parenthesized Peak values expose the retained dormant anchors; Table~\ref{tab:component_ablation}
further separates captioned and CLIP-resident states.
Across Tier~1/2, JITOMA uses only 1--2 active objects at query time
(peak $\leq4$) while retaining 284--512 dormant anchors; Tier~3 uses six
active objects with a peak of twelve.
Thus, our claim concerns bounded semantic materialization rather than
negligible total storage.
Despite invoking on-demand dense captioning, JITOMA maintains
$0.25$--$0.28$\,s/frame, at or below Clio's $0.27$--$0.46$\,s/frame across
all settings.
This shows that JIT allocation preserves a persistent visual memory while
avoiding global semantic expansion.

\subsection{Component Ablation}
\label{sec:ablation}

We ablate the two mechanisms that define JITOMA's just-in-time growth process:
\emph{task-aware observation admission} at the perception frontend and
\emph{on-demand semantic activation} at the memory backend.
The former determines which observations are committed to memory, whereas
the latter delays expensive semantic expansion until an object becomes
relevant to the current task.
We conduct the study on Tier~2 of JITOMA-Bench, which evaluates continual
grounding under a stream of task switches with persistent scene memory.
The default hyperparameters are used throughout this analysis, and their
sensitivity is further studied in Appendix~\ref{app:hyperparameter_sensitivity}.

\paragraph{Ablation design.}
We decompose JITOMA along its two AOT--JIT transitions: observation admission
at the frontend and semantic expansion at the backend.
At the frontend, AOT admits observations independently of the task,
whereas JIT performs task-guided admission; at the backend, AOT expands
all objects proactively, whereas JIT activates semantics on demand.
Variants A, B, C, and E span the resulting $2\times2$ design, with E denoting
the full JITOMA.
Variant D$^\dagger$ is a task-independent control that matches E's
admission budget while selecting observations without task,
isolating task relevance from sparsity.
Results are averaged over the three Tier~2 scenes and summarized in
Table~\ref{tab:component_ablation}.

\begin{table}[ht]
    \centering
    \small
    \caption{
    Component ablation on Tier~2 of JITOMA-Bench.
    AOT/JIT denote task-independent/task-guided frontend admission and
    proactive/on-demand backend expansion.
    $^\dagger$ matches JITOMA's per-frame admission budget.
    Cap.\ and CLIP denote captioned and CLIP-resident objects.
    }
    \label{tab:component_ablation}

    \setlength{\tabcolsep}{4.0pt}
    \renewcommand{\arraystretch}{1.05}

    \resizebox{\columnwidth}{!}{%
    \begin{tabular}{lcccccccccc}
        \toprule
        & \multicolumn{2}{c}{Configuration}
        & \multicolumn{3}{c}{Accuracy $\uparrow$}
        & \multicolumn{5}{c}{Efficiency $\downarrow$} \\
        \cmidrule(lr){2-3}
        \cmidrule(lr){4-6}
        \cmidrule(lr){7-11}
        
        Variant
        & Frontend
        & Backend
        & IoU
        & mR@1
        & mR@3
        & Objs$_{\mathrm{Cap}}$
        & Objs$_{\mathrm{CLIP}}$
        & Peak$_{\mathrm{Cap}}$
        & Peak$_{\mathrm{CLIP}}$
        & TPF \\
        \midrule

        A
        & AOT & AOT
        & 12.8 & 31.0 & 45.9
        & 355.0 & 246.9
        & 512.0 & 387.0
        & 0.460 \\

        B
        & JIT & AOT
        & 22.9 & 52.4 & 64.7
        & 223.4 & 170.8
        & 421.0 & 331.3
        & 0.436 \\

        C
        & AOT & JIT
        & 13.5 & 30.4 & 45.9
        & 31.4 & 247.1
        & 49.7 & 387.0
        & 0.336 \\

        D$^\dagger$
        & AOT$^\dagger$ & JIT
        & 13.7 & 31.9 & 42.6
        & 31.0 & 208.1
        & 51.0 & 368.3
        & 0.311 \\

        \rowcolor{black!15}
        \textbf{E (Ours)}
        & \textbf{JIT} & \textbf{JIT}
        & \textbf{24.3} & \textbf{57.5} & \textbf{67.1}
        & \textbf{28.5} & \textbf{170.5}
        & \textbf{47.0} & \textbf{330.7}
        & \textbf{0.282} \\

        \bottomrule
    \end{tabular}%
    }
\end{table}

\paragraph{Task-aware perception limits memory accumulation.}
Comparing A and B isolates the frontend JIT transition.
Task-guided admission reduces the average CLIP-resident object count from
246.9 to 170.8 and its peak from 387.0 to 331.3, while lowering TPF from
0.460 to 0.436\,s/frame and improving IoU from 12.8 to 22.9 and mR@1 from
31.0 to 52.4.
Thus, frontend JIT reduces the visual state committed to persistent memory
before semantic processing.

\paragraph{On-demand activation limits semantic computation.}
Comparing A and C isolates the backend transition.
With nearly unchanged CLIP-resident memory, JIT activation reduces the
average number of captioned objects from 355.0 to 31.4 and the peak from
512.0 to 49.7, while lowering TPF from 0.460 to 0.336\,s/frame.
Thus, backend JIT primarily avoids proactive semantic materialization rather
than eliminating the dormant visual memory itself.

\paragraph{Complementary frontend and backend resource allocation.}
The two JIT transitions reduce different resource costs.
Frontend task-aware admission controls \emph{what enters memory}, limiting
persistent visual-state growth and downstream processing.
Backend on-demand activation controls \emph{when semantic computation is
incurred}, reducing captioned objects from 355.0 to 31.4 and TPF from
0.460 to 0.336\,s/frame while leaving CLIP-resident memory nearly unchanged.
Together, JITOMA limits persistent memory growth at the frontend and semantic
compute at the backend.

\subsection{Qualitative Analysis of Just-In-Time Memory Growth}
\label{sec:case_study}

\begin{figure*}[ht]
    \centering
    \includegraphics[width=\textwidth]{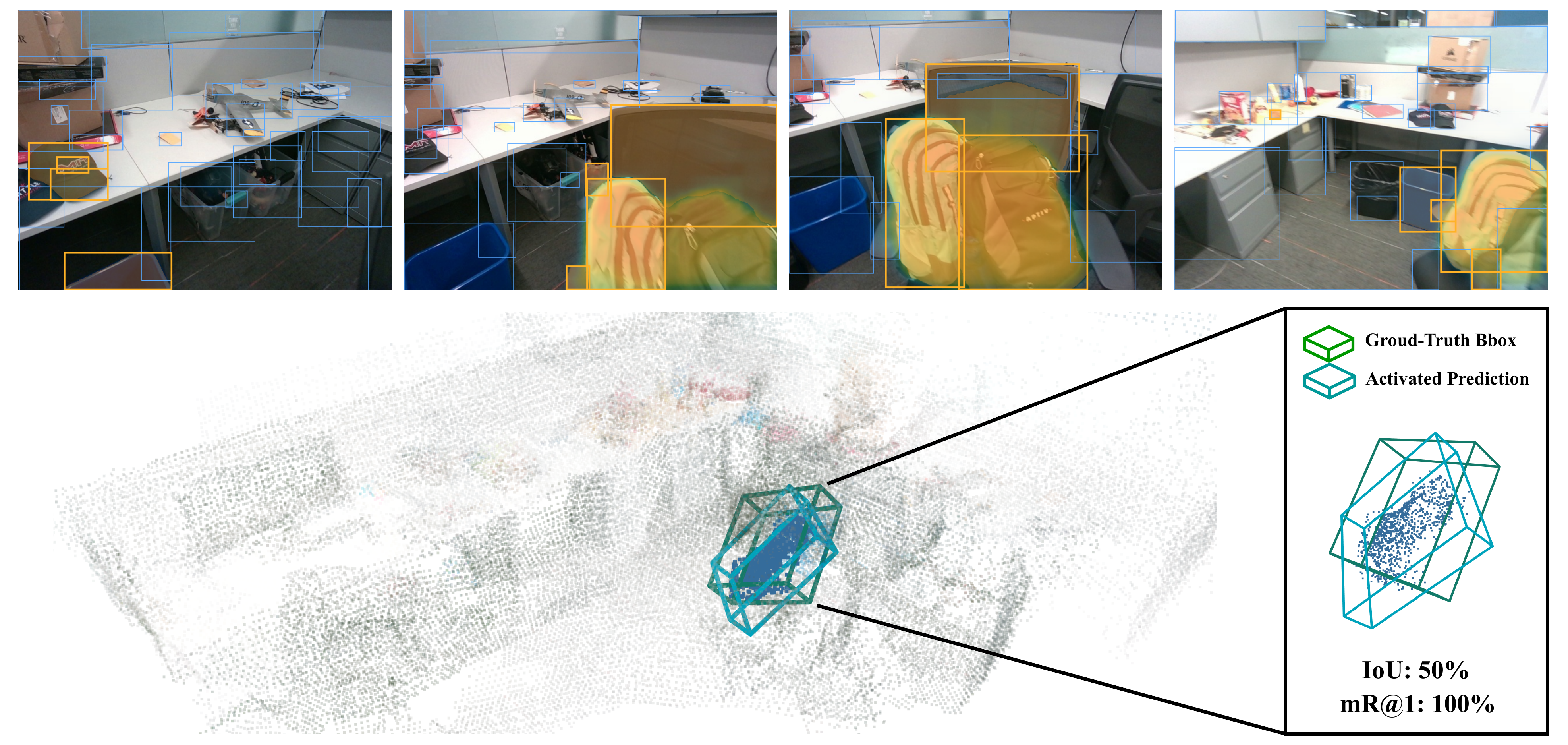}
    \caption{\textbf{Qualitative example of just-in-time memory growth.}
The example illustrates JITOMA's task-guided perception and on-demand
semantic activation under the ``Get backpacks'' command.
\textbf{Top:} During frontend perception, blue boxes denote tracked objects
that are not admitted because they provide neither task relevance nor novel
visual evidence. Orange boxes denote newly admitted observations, where
heatmap-highlighted regions correspond to task-relevant objects and
non-highlighted regions correspond to novel but task-irrelevant objects.
\textbf{Bottom:} JITOMA activates only the task-relevant anchor from the
dormant memory foundation. The activated prediction (cyan) is compared with
the ground-truth bounding box (green), achieving IoU=50\% and mR@1=100\%.
}
    \label{fig:case_study}
\end{figure*}

Fig.~\ref{fig:case_study} illustrates how JITOMA allocates perception and
memory resources under the ``Get backpacks'' command.
During frontend perception, previously observed but irrelevant objects
(blue boxes) are suppressed without forming new memory hypotheses, while
novel observations (orange boxes) are selectively admitted according to
task relevance and visual novelty.
After activation, JITOMA expands only the task-relevant dormant anchors,
producing a compact semantic representation whose activated prediction
aligns with the ground-truth object.
This example demonstrates that JITOMA separates persistent geometric memory
from task-specific semantic growth, enabling detailed semantic processing only when
required by the current task.

%% file: Tables/Table1_main_result.tex
\definecolor{accBg}{HTML}{FFF0F5} 
\definecolor{effBg}{HTML}{FFFFF0} 

\begin{table*}[ht]
\centering
\renewcommand{\arraystretch}{1.15} 
\setlength{\tabcolsep}{4pt}
\caption{\textbf{Evaluation on JITOMA-Bench.}
We compare JITOMA with sota baselines across all three evaluation
tiers, measuring grounding accuracy and efficiency.
IoU, mR@1, and mR@3 measure grounding accuracy, while Objs, Peak, and TPF
measure query-time active semantic size, peak active semantic size, and
per-frame latency.
For JITOMA, Peak is reported as active(dormant), explicitly exposing the
retained dormant-anchor state.
For Tier~3, baseline accuracy is reported as \textbf{Vanilla / Augmented},
where Vanilla uses each method's original task interpretation pipeline and
Augmented replaces task parsing with our LLM-based parser to isolate retrieval
capability from intent parsing.
}

\label{tab:main_results}
\resizebox{\textwidth}{!}{%
\begin{tabular}{c l 
    >{\columncolor{accBg}}c >{\columncolor{accBg}}c >{\columncolor{accBg}}c 
    >{\columncolor{effBg}}c >{\columncolor{effBg}}c >{\columncolor{effBg}}c 
    >{\columncolor{accBg}}c >{\columncolor{accBg}}c >{\columncolor{accBg}}c 
    >{\columncolor{effBg}}c >{\columncolor{effBg}}c >{\columncolor{effBg}}c 
    >{\columncolor{accBg}}c >{\columncolor{accBg}}c >{\columncolor{accBg}}c 
    >{\columncolor{effBg}}c >{\columncolor{effBg}}c >{\columncolor{effBg}}c}
\toprule
\multirow{3}{*}{\textbf{Scene}} & \multirow{3}{*}{\textbf{Method}} & \multicolumn{6}{c}{\textbf{Tier 1: Foundational Single-Goal}} & \multicolumn{6}{c}{\textbf{Tier 2: Long-Horizon Temporal}} & \multicolumn{6}{c}{\textbf{Tier 3: Complex Intent}} \\
\cmidrule(lr){3-8} \cmidrule(lr){9-14} \cmidrule(lr){15-20} 

& & \multicolumn{3}{c}{\cellcolor{accBg}\textit{Accuracy $\uparrow$}} & \multicolumn{3}{c}{\cellcolor{effBg}\textit{Efficiency}} & \multicolumn{3}{c}{\cellcolor{accBg}\textit{Accuracy $\uparrow$}} & \multicolumn{3}{c}{\cellcolor{effBg}\textit{Efficiency}} & \multicolumn{3}{c}{\cellcolor{accBg}\textit{Accuracy $\uparrow$}} & \multicolumn{3}{c}{\cellcolor{effBg}\textit{Efficiency}} \\
\cmidrule(lr){3-5} \cmidrule(lr){6-8} \cmidrule(lr){9-11} \cmidrule(lr){12-14} \cmidrule(lr){15-17} \cmidrule(lr){18-20}

& & IoU & mR@1 & mR@3 & Objs $\downarrow$ & Peak $\downarrow$ & TPF $\downarrow$ 
& IoU & mR@1 & mR@3 & Objs $\downarrow$ & Peak $\downarrow$ & TPF $\downarrow$ 
& IoU & mR@1 & mR@3 & Objs $\downarrow$ & Peak $\downarrow$ & TPF $\downarrow$ \\
\midrule

\multirow{4}{*}{\rotatebox{90}{\textbf{Apartment}}} 
& ConceptGraphs 
& 10.9 & 20.5 & 28.2 & 444 & 1479 & 5.22 
& 17.6 & 33.3 & \textbf{52.8} & 452 & 1476 & 6.03  
& 2.9 / 14.3 & 7.4 / 25.9 & 7.4 / 35.2 & 451 & 1494 & 5.26 \\ 

& ReasoningGraph
& 9.7 & 25.6 & 26.9 & 73 & 488 & 2.91
& 12.6 & 33.3 & 33.3 & 105 & 488 & 2.91
& 4.4 / 6.4 & 11.1 / 16.7 & 11.1 / 16.7 & 73 & 488 & 2.91 \\

& Clio 
& 9.5 & 19.2 & 49.3 & 29 & 717 & 0.27 
& 10.9 & 25.0 & 27.8 & 4 & 637 & 0.35  
& 5.6 / 9.5 & 12.5 / 21.5 & 16.7 / 25.7 & 30 & 704   & 0.27 \\

& JITOMA (Ours) 
& \textbf{16.2} & \textbf{32.1} & \textbf{51.3} & 2 & 4(488) & \textbf{0.25 }
& \textbf{21.4} & \textbf{47.2} & \textbf{52.8} & 2 & 4(494) & \textbf{0.26}
& \textbf{14.8} & \textbf{27.0} & \textbf{35.4} & 6 & 12(467) & \textbf{0.26} \\ 

\midrule

\multirow{4}{*}{\rotatebox{90}{\textbf{Office}}} 
& ConceptGraphs 
& 13.5 & 31.8 & 37.9 & 926 & 5793 & 11.10  
& 19.0 & 40.5 & 45.2 & 924 & 5796 & 12.58  
& 1.8 / 13.3 & 0.0 / 35.7 & 4.8 / 47.6 & 931 & 5797 & 12.13 \\

& ReasoningGraph
& 8.4 & 21.2 & 36.4 & 35 & 605 & 3.04
& 17.3 & 42.9 & 45.2 & 102 & 605 & 3.04
& 10.2 / 9.2 & 26.2 / 23.8 & 26.2 / 47.6 & 35 & 605 & 3.04 \\

& Clio 
& 17.2 & 39.4 & 56.0 & 19 & 723 & 0.28  
& 17.7 & 38.1 & 47.6 & 5 & 678 & 0.42  
& 1.7 / 16.5 & 5.6 / 40.7 & 5.6 / 50.0 & 19 & 619 & 0.29 \\

& JITOMA (Ours) 
& \textbf{20.4} & \textbf{53.0} & \textbf{60.6} & 1 & 3(510) & \textbf{0.25}
& \textbf{29.1} & \textbf{64.3} & \textbf{73.8} & 1 & 4(512) & \textbf{0.26} 
& \textbf{19.8} & \textbf{50.7} & \textbf{58.0} & 6 & 12(512) & \textbf{0.26} \\

\midrule

\multirow{4}{*}{\rotatebox{90}{\textbf{Cubicle}}} 
& ConceptGraphs 
& 11.2 & 24.1 & 38.9 & 235 & 758 & 5.18 
& 11.3 & 22.2 & 63.0 & 224 & 745 & 5.37  
& 3.7 / 6.4  & 4.4 / 5.8 & 15.9 / 31.9 & 237 & 766 & 5.32 \\

& ReasoningGraph
& 8.4 & 18.5 & 24.1 & 89 & 260 & 2.92
& 12.9 & 22.2 & 29.6 & 124 & 260 & 2.92
& 3.6 / 9.3 & 10.1 / 18.8 & 15.9 / 29.0 & 89 & 260 & 2.92 \\

& Clio
& 19.3 & 48.1 & 59.3 & 41 & 370 & 0.33
& \textbf{23.8} & \textbf{66.7} & \textbf{74.1} & 5 & 408 & 0.46  
& 7.6 / 17.3 & 18.2 / 44.4 & 21.2 / \textbf{57.6} & 44 & 460 & 0.33 \\

& JITOMA (Ours) 
& \textbf{21.9} & \textbf{55.6} & \textbf{66.7} & 2 & 4(284) & \textbf{0.26} 
& 22.4 & \textbf{66.7} & \textbf{74.1} & 2 & 4(289) & \textbf{0.28} 
& \textbf{21.6} & \textbf{45.2} & \textbf{57.6} & 6 & 12(288) & \textbf{0.28} \\

\bottomrule
\end{tabular}
}
\end{table*}

%% file: Sections/7_conclusion.tex
\section{Conclusion}
\label{sec:conclusion}

We introduced \textbf{JITOMA}, a just-in-time scene graph framework that
shifts 3D memory construction from exhaustive Ahead-of-Time accumulation to
task-conditioned on-demand growth. By maintaining lightweight dormant anchors
and activating semantic structures only when required, JITOMA achieves
efficient embodied scene understanding while preserving grounding capability
under increasingly complex tasks. Through JITOMA-Bench and extensive
experiments, we demonstrate that controlling not only \emph{what} a scene
graph represents but also \emph{when and where} its semantic structure is
instantiated is essential for scalable robotic perception.

%% file: Sections/Appendix/3_bench.tex
\subsection{JITOMA-Bench Details}
\label{app:jitoma_bench_details}

\subsubsection{Dataset Statistics}
\label{app:jitoma_bench_statistics}

\begin{figure*}[ht]
    \centering
    \includegraphics[width=\textwidth]{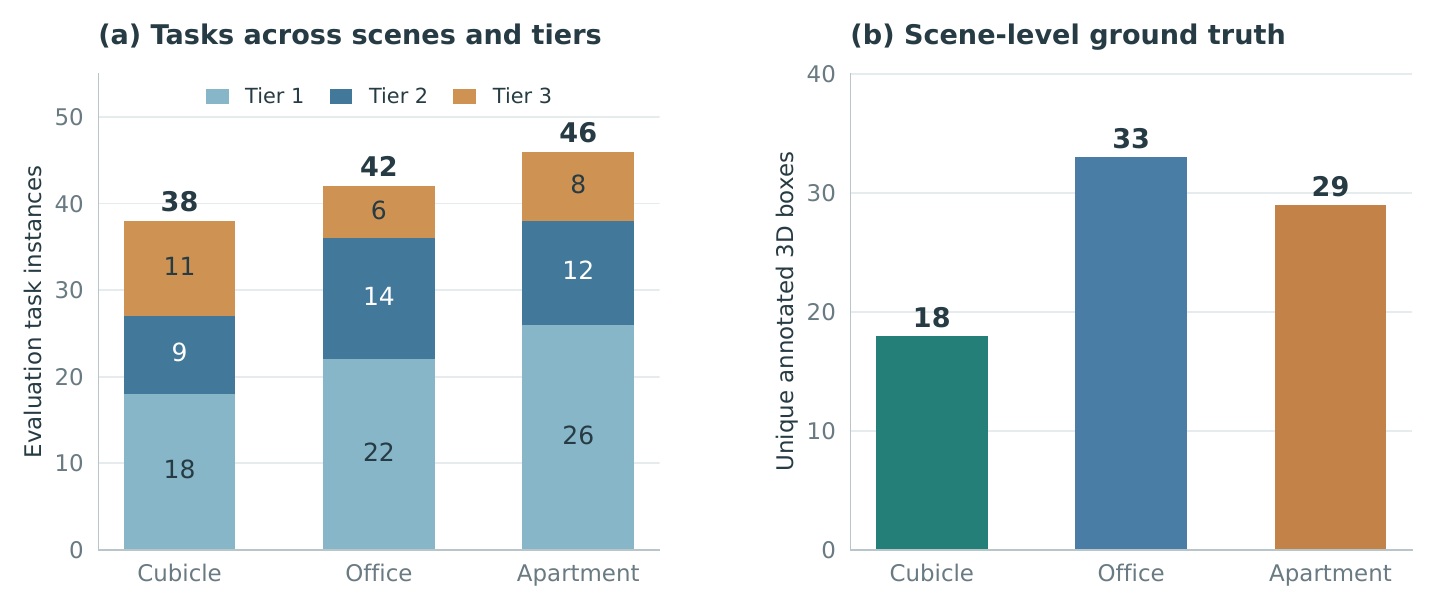}
    \caption{\textbf{Composition of JITOMA-Bench.}
    \textbf{(a)} Number of evaluation tasks in each scene, decomposed by
    Tier~1, Tier~2, and Tier~3.
    \textbf{(b)} Number of unique annotated 3D ground-truth boxes per scene.
    JITOMA-Bench contains 126 task instances and 80 unique GT boxes in total;
    Tier~2 reuses Tier~1 tasks and ground truth under a continual evaluation
    protocol.}
    \label{fig:bench_composition}
\end{figure*}

JITOMA-Bench contains 126 evaluation task instances across the Cubicle,
Office, and Apartment scenes.
As summarized in Fig.~\ref{fig:bench_composition}, the three scenes contain
38, 42, and 46 task instances, respectively, spanning all three evaluation
tiers.
Across the benchmark, these tasks refer to 80 unique annotated 3D
ground-truth boxes.
Tier~2 intentionally reuses Tier~1 objects within continuous task streams;
therefore, repeated task--object references across tiers should not be
interpreted as additional unique ground-truth instances.

\begin{figure*}[ht]
    \centering
    \includegraphics[width=\textwidth]{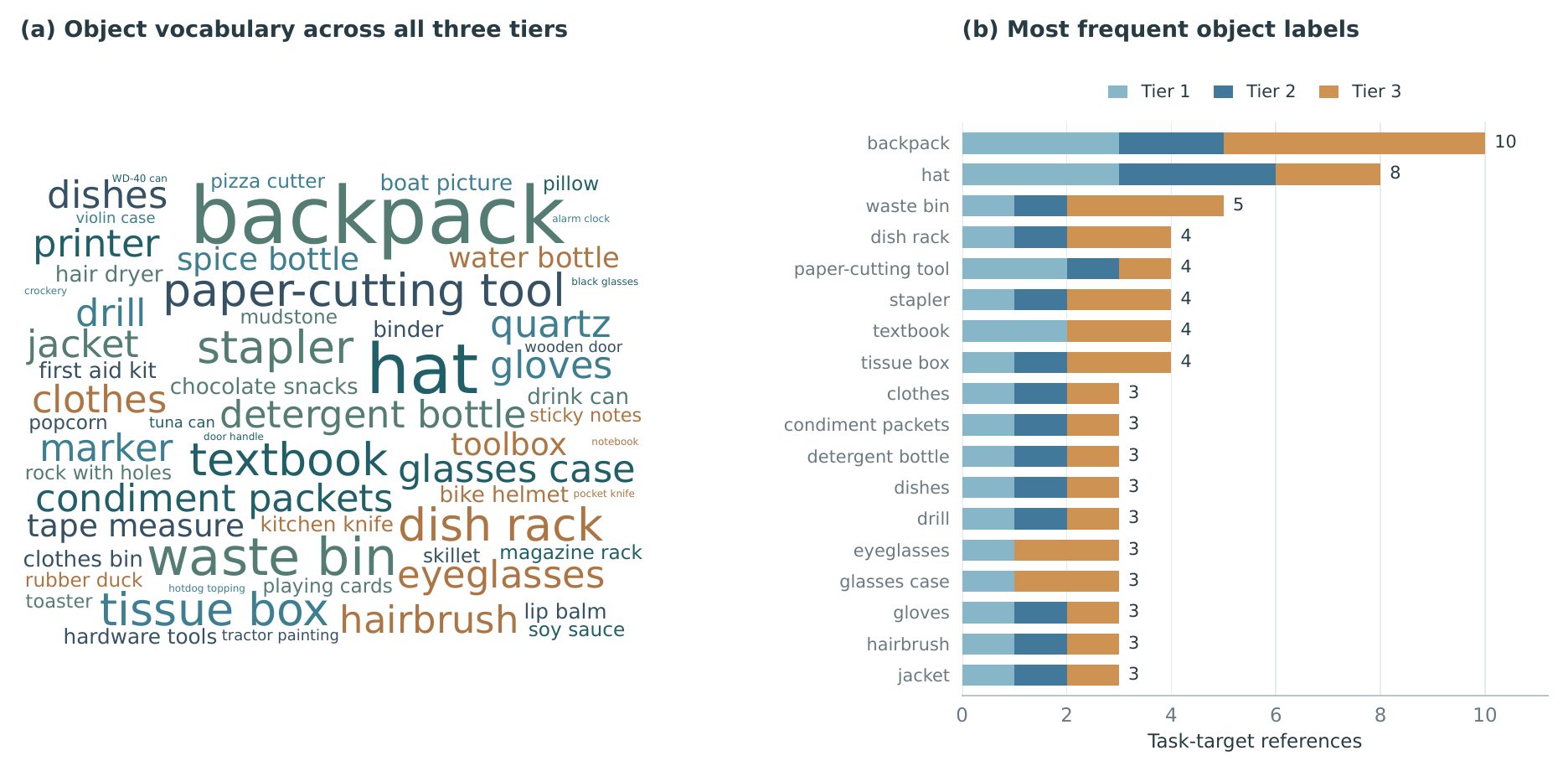}
    \caption{\textbf{Object vocabulary and target frequency in JITOMA-Bench.}
    \textbf{(a)} Object vocabulary aggregated across all three tiers.
    \textbf{(b)} Most frequently referenced object labels, decomposed by tier.
    The benchmark contains 156 task-target references in total
    (66/35/55 for Tier~1/2/3).
    Counts include intentional Tier~2 reuse; display aliases merge selected
    surface forms rather than ground-truth instances.}
    \label{fig:bench_object_frequency}
\end{figure*}

The benchmark also covers a broad object vocabulary rather than repeatedly
querying a small set of categories.
Fig.~\ref{fig:bench_object_frequency} aggregates all task-target references
across the three tiers, yielding 156 references in total:
66 from Tier~1, 35 from Tier~2, and 55 from Tier~3.
Frequently referenced objects such as \emph{backpack} and \emph{hat} occur
across multiple tasks and tiers, while most object labels appear only a small
number of times.
This long-tailed distribution reflects the open-vocabulary nature of the
underlying scenes and reduces dependence on a small set of repeatedly queried
categories.
For visualization, selected surface-form aliases are merged, while the
underlying ground-truth instances remain unchanged.

\begin{figure*}[ht]
    \centering
    \includegraphics[width=\textwidth]{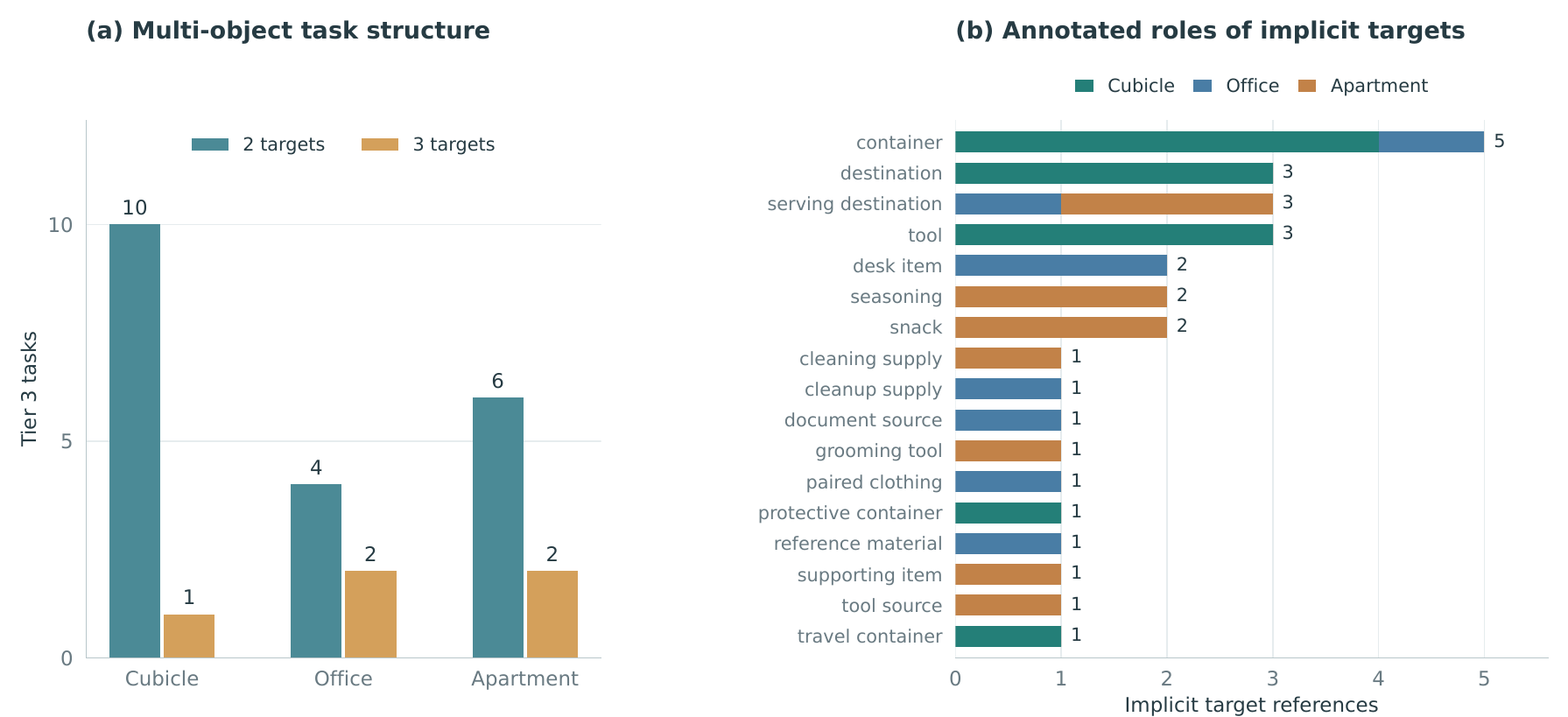}
    \caption{\textbf{Structure of Tier~3 complex tasks.}
    \textbf{(a)} Distribution of two- and three-target instructions across
    the three scenes.
    \textbf{(b)} Functional roles assigned to latent targets.
    Tier~3 contains 25 tasks with 25 primary and 30 latent target references.
    Role names are preserved directly from the annotations without post-hoc
    category merging.}
    \label{fig:bench_complex_structure}
\end{figure*}

Tier~3 further increases compositional complexity by requiring each
instruction to ground multiple objects.
As shown in Fig.~\ref{fig:bench_complex_structure}, the 25 Tier~3 tasks
contain 25 primary and 30 latent target references.
Twenty tasks involve two targets and five involve three, with all three scenes
contributing multi-object instructions.
The latent targets span diverse functional roles, including containers,
destinations, tools, serving destinations, cleaning supplies, reference
materials, and supporting objects.
We retain the role descriptions assigned during annotation rather than
post-hoc merging them into broader categories, preserving the functional
diversity of the original instructions.

\subsubsection{Task Annotation Protocol}
\label{app:jitoma_bench_annotation}

Tier~1 provides the base set of explicit single-object grounding tasks.
Each task is evaluated independently using the corresponding Clio
ground-truth object annotations.

\begin{figure*}[ht]
    \centering
    \includegraphics[width=\textwidth]{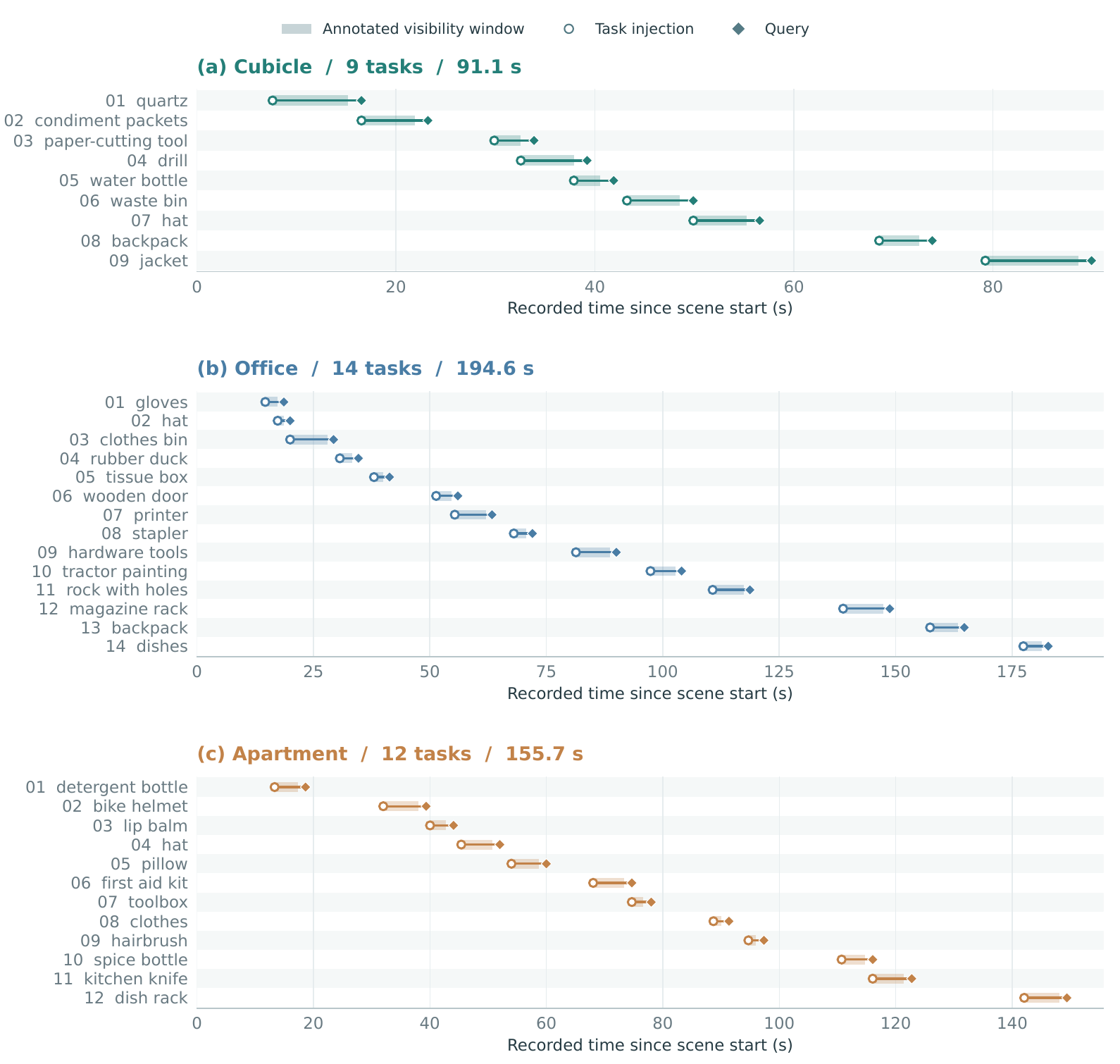}
    \caption{\textbf{Tier~2 temporal task annotations.}
    Annotated target-visibility intervals, task-injection times, and query
    times for the Cubicle, Office, and Apartment trajectories.
    Tasks are injected at the first-visible frame and evaluated ten frames
    after the annotated last-visible frame.
    Time is shown using the recorded RGB-D timestamps; the visibility windows
    are manual annotations and should not be interpreted as inferred runtime
    concurrency.}
    \label{fig:bench_timelines}
\end{figure*}

For Tier~2, we manually annotate the temporal visibility interval of each
target along the original RGB-D trajectory.
A task becomes active at its first annotated visible frame and is queried
ten frames after its last-visible frame.
Fig.~\ref{fig:bench_timelines} visualizes these intervals over recorded
trajectory time.
The Cubicle, Office, and Apartment sequences contain 9, 14, and 12 Tier~2
tasks over 91.1\,s, 194.6\,s, and 155.7\,s of recorded RGB-D data,
respectively.
These task windows are annotation intervals rather than inferred runtime
concurrency; the ten-frame query offset therefore also does not correspond
to a fixed delay in seconds.
This protocol preserves a persistent scene memory throughout the complete
trajectory and evaluates the system under repeated task transitions without
resetting the map between queries.

Tier~3 is constructed from the same scene annotations but replaces atomic
object queries with multi-object instructions containing both explicit
primary objects and execution-implied latent objects.
Candidate instructions are generated from Clio ground-truth labels and then
manually assigned primary and latent targets.
Three PhD researchers independently review each candidate, and an instruction
is discarded when at least two reviewers identify ambiguity, infeasibility,
or unnecessary latent targets.
The resulting 25 tasks contain the multi-object and functional-role
distributions summarized in Fig.~\ref{fig:bench_complex_structure}.

%% file: Sections/Appendix/0_moti_exp.tex
\subsection{Controlled Study of Perceptual Saturation}
\label{supp:perceptual_saturation}

\subsubsection{Prompt and Graph Construction}
\label{supp:prompt_aot_motivation}
\label{supp:prompt_jit_motivation}

To isolate the effect of task-irrelevant semantic context, the AOT and JIT
conditions use exactly the same task instruction, planner, prompt template,
output format, and decoding configuration.
The only difference is the set of scene-graph entities exposed to the planner.
All task-relevant entities retain identical identifiers, object names, and
geometric attributes across conditions.
The JIT condition contains only the entities required for task execution,
whereas the AOT condition additionally includes task-irrelevant objects from
the same scene.

For the qualitative example in Sec.~\ref{sec:preliminary_findings}, the task is
``pack my eyeglasses for travel'' in the Cubicle scene.
The task-relevant condition contains the three required entities
(\emph{eyeglasses}, \emph{glasses case}, and \emph{backpacks}), while the
full-graph condition contains the same three entries together with 15
additional distractors.

\paragraph{AOT full graph.}
\begin{tcolorbox}[
    colback=orange!5!white,
    colframe=orange!60!black,
    title=Ahead-of-Time (AOT) Full Graph,
    fonttitle=\bfseries,
    arc=1mm,
    breakable
]
\footnotesize
\textbf{Task:} pack my eyeglasses for travel \\
\textbf{Scene:} cubicle \\

\vspace{0.5em}
\textbf{Scene graph:} \\
{\scriptsize\ttfamily
- id="full\_001", object="condiment packets" \\
\ \ $\hookrightarrow$ coords=[center=(-0.83, 1.62, -0.76), extents=(0.27, 0.19, 0.01)] \\
- id="full\_002", object="drink cans" \\
\ \ $\hookrightarrow$ coords=[center=(-0.41, 4.03, -0.83), extents=(0.14, 0.15, 0.09)] \\
- id="full\_003", object="eyeglasses" \\
\ \ $\hookrightarrow$ coords=[center=(-0.83, 1.26, -0.70), extents=(0.10, 0.03, 0.02)] \\
- id="full\_004", object="glasses case" \\
\ \ $\hookrightarrow$ coords=[center=(-0.82, 1.27, -0.72), extents=(0.14, 0.04, 0.02)] \\
- id="full\_005", object="grey jacket" \\
\ \ $\hookrightarrow$ coords=[center=(-0.02, 0.72, -0.71), extents=(0.72, 0.26, 0.15)] \\
- id="full\_006", object="my silver water bottle" \\
\ \ $\hookrightarrow$ coords=[center=(-0.52, 4.01, -0.77), extents=(0.23, 0.05, 0.04)] \\
- id="full\_007", object="notebooks" \\
\ \ $\hookrightarrow$ coords=[center=(-0.26, 3.70, -0.86), extents=(0.66, 0.34, 0.03)] \\
- id="full\_008", object="mudstone rock" \\
\ \ $\hookrightarrow$ coords=[center=(-1.00, 0.11, -0.63), extents=(0.13, 0.07, 0.06)] \\
- id="full\_009", object="tool to cut paper" \\
\ \ $\hookrightarrow$ coords=[center=(-0.77, 2.86, -0.81), extents=(0.30, 0.13, 0.03)] \\
- id="full\_010", object="sticky notes" \\
\ \ $\hookrightarrow$ coords=[center=(-1.22, 1.32, -0.65), extents=(0.18, 0.18, 0.01)] \\
- id="full\_011", object="textbooks" \\
\ \ $\hookrightarrow$ coords=[center=(-1.04, 2.08, -0.73), extents=(0.60, 0.41, 0.08)] \\
- id="full\_012", object="waste bins" \\
\ \ $\hookrightarrow$ coords=[center=(0.17, 3.27, -1.36), extents=(0.83, 0.55, 0.26)] \\
- id="full\_013", object="hats" \\
\ \ $\hookrightarrow$ coords=[center=(0.31, 3.57, -0.82), extents=(0.34, 0.22, 0.08)] \\
- id="full\_014", object="backpacks" \\
\ \ $\hookrightarrow$ coords=[center=(0.75, 2.56, -0.82), extents=(0.61, 0.59, 0.21)] \\
- id="full\_015", object="red crockery" \\
\ \ $\hookrightarrow$ coords=[center=(-0.87, 3.92, -0.84), extents=(0.47, 0.27, 0.09)] \\
- id="full\_016", object="hardware drill" \\
\ \ $\hookrightarrow$ coords=[center=(-0.86, 3.58, -0.83), extents=(0.20, 0.17, 0.04)] \\
- id="full\_017", object="quartz rock" \\
\ \ $\hookrightarrow$ coords=[center=(-0.91, 0.09, -0.64), extents=(0.10, 0.08, 0.04)] \\
- id="full\_018", object="tape measure" \\
\ \ $\hookrightarrow$ coords=[center=(-0.72, 3.61, -0.83), extents=(0.09, 0.08, 0.06)]
}
\end{tcolorbox}

\paragraph{JIT task-relevant graph.}
\begin{tcolorbox}[
    colback=teal!5!white,
    colframe=teal!60!black,
    title=Just-In-Time (JIT) Task-Relevant Graph,
    fonttitle=\bfseries,
    arc=1mm,
    breakable
]
\footnotesize
\textbf{Task:} pack my eyeglasses for travel \\
\textbf{Scene:} cubicle \\

\vspace{0.5em}
\textbf{Scene graph:} \\
{\scriptsize\ttfamily
- id="full\_003", object="eyeglasses" \\
\ \ $\hookrightarrow$ coords=[center=(-0.83, 1.26, -0.70), extents=(0.10, 0.03, 0.02)] \\
- id="full\_004", object="glasses case" \\
\ \ $\hookrightarrow$ coords=[center=(-0.82, 1.27, -0.72), extents=(0.14, 0.04, 0.02)] \\
- id="full\_014", object="backpacks" \\
\ \ $\hookrightarrow$ coords=[center=(0.75, 2.56, -0.82), extents=(0.61, 0.59, 0.21)]
}
\end{tcolorbox}

\paragraph{Shared planner prompt.}
The following prompt is used unchanged for both graph conditions:

\begin{tcolorbox}[
    colback=gray!5!white,
    colframe=gray!60!black,
    title=Shared Task-Planning Prompt,
    fonttitle=\bfseries,
    arc=1mm,
    breakable
]
\footnotesize

\textbf{Input Prompt:} \\
Return the answer in exactly this style, with 3 or 4 numbered quoted steps
and nothing else:

\begin{enumerate}[label=\arabic*., itemsep=1pt, topsep=2pt, leftmargin=*]
    \item \texttt{"Action using object [center=(x, y, z)]",}
    \item \texttt{"Next action using object [center=(x, y, z)]",}
    \item \texttt{"Next action using object [center=(x, y, z)]",}
    \item \texttt{"Final action using object [center=(x, y, z)]"}
\end{enumerate}

\vspace{0.4em}
\textbf{Rules:}
\begin{enumerate}[leftmargin=1.5em, itemsep=2pt, topsep=2pt]
    \item Mention all objects that are required to complete the task.
    \item If a destination, container, support, source, or tool is needed and
    exists in the graph, include it.
    \item Every mentioned object must include its center coordinate exactly
    once in that step.
    \item Use short imperative actions. Do not explain.
\end{enumerate}

\end{tcolorbox}

\subsubsection{Controlled Distractor Analysis}
\label{supp:perceptual_saturation_controlled}

To examine whether the qualitative failure mode in
Sec.~\ref{sec:preliminary_findings} extends beyond a single example, we
conduct a controlled distractor study over the Tier~3 tasks of JITOMA-Bench.
For each task, all task-required entities and their attributes are held fixed,
while $k\in\{0,5,10,15\}$ task-irrelevant entities from the same scene are
added to the planner input.
All conditions use the same planner, prompt template, and output format
described in Appendix~\ref{supp:prompt_aot_motivation}; thus, the manipulated
factor is the amount of irrelevant semantic context presented alongside the
task-relevant graph.
Across the three scenes and repeated distractor configurations, the study
contains 1,194 planner requests, all of which complete without interface
errors or output truncation.

\paragraph{Evaluation metric.}
We measure \emph{target coverage}, defined as the fraction of annotated
task-required objects explicitly referenced by the generated plan.
We report coverage over all required targets and separately over
\emph{latent targets}, i.e., objects not explicitly named in the command but
required for task execution.
The latter provides a more demanding measure of whether the planner recovers
implicit functional dependencies such as containers, destinations, or tools.

\begin{table}[ht]
    \centering
    \small
    \setlength{\tabcolsep}{7pt}
    \caption{
    Controlled distractor analysis on Tier~3 of JITOMA-Bench.
    Task-relevant graph content is held fixed while $k$ task-irrelevant
    entities are added.
    Results are averaged across the three scenes.
    }
    \label{tab:perceptual_saturation_controlled}
    \begin{tabular}{ccc}
        \toprule
        Distractors $k$
        & All-target coverage (\%)
        & Latent-target coverage (\%) \\
        \midrule
        0  & 82.31 & 68.10 \\
        5  & 79.26 & 62.08 \\
        10 & 78.23 & 60.97 \\
        15 & 75.24 & 56.30 \\
        \midrule
        $\Delta(15-0)$
        & $-7.07$
        & $-11.81$ \\
        \bottomrule
    \end{tabular}
\end{table}

\paragraph{Effect of irrelevant semantic load.}
As shown in Table~\ref{tab:perceptual_saturation_controlled}, target coverage
decreases as additional task-irrelevant entities are introduced.
From $k=0$ to $k=15$, all-target coverage drops from 82.31\% to 75.24\%,
while latent-target coverage decreases more substantially from 68.10\% to
56.30\%.
A paired bootstrap comparison between these two conditions yields 95\%
confidence intervals of $[-13.89,-1.42]$ percentage points for all-target
coverage and $[-23.21,-2.12]$ for latent-target coverage.
The larger degradation on latent targets indicates that irrelevant semantic
context particularly interferes with recovering objects implied by the
execution chain rather than explicitly stated in the command.

This behavior is not explained by loss of the primary task target:
primary-target coverage remains nearly unchanged, from 98.52\% at $k=0$ to
98.11\% at $k=15$.
The latent-target degradation is also observed across all three scenes:
coverage decreases from 100.00\% to 93.33\% in Cubicle,
67.22\% to 47.22\% in Office, and 37.08\% to 28.33\% in Apartment.
Qualitative inspection reveals recurring failures such as omitting a required
container or destination, terminating a multi-step plan prematurely, or
introducing operations on irrelevant objects.

\paragraph{Interpretation.}
The effect is not deterministic for every individual task.
Between $k=0$ and $k=15$, latent-target coverage decreases for 9 of the
25 Tier~3 tasks, remains unchanged for 14, and improves for 2.
We therefore interpret perceptual saturation as a population-level tendency
under increasing irrelevant semantic load rather than a universal failure
mode.
Together with the qualitative example in
Sec.~\ref{sec:preliminary_findings}, these results provide controlled evidence
that exposing a planner to additional task-irrelevant scene semantics can
reduce recovery of task-required information, with the strongest effect on
latent functional dependencies.

%% file: Sections/Appendix/1_hyper.tex
\subsection{Deployment Details}
\label{app:deployment_details}

\subsubsection{Hardware and Software Environment}
\label{app:hardware}

Each experimental run uses a single NVIDIA RTX A6000 GPU with 48\,GB VRAM
and is executed on Ubuntu 22.04.
The main JITOMA pipeline and the local Qwen service share the same GPU but
run in separate software environments.
The main pipeline uses Python 3.10, PyTorch 2.0.1 with CUDA 11.8, and
Transformers 4.41.2, while the local LLM service uses Python 3.10,
PyTorch 2.4.0 with CUDA 12.1, and Transformers 5.6.0.dev0.
Qwen3.5-9B is served locally in BF16 precision.

\paragraph{Runtime measurement.}
TPF is computed as the total wall-clock time required to process a complete
scene replay divided by the number of processed RGB-D frames.
The measurement covers the full online mapping pipeline, including perception,
memory updates, task-conditioned processing, and on-demand semantic operations
such as DAM captioning whenever they are triggered.
The same timing protocol is applied to all methods.

\subsubsection{Hyperparameter Settings}
\label{app:hyperparameter_settings}

JITOMA contains a small set of hyperparameters governing three stages of
the just-in-time growth process: task-conditioned observation admission,
memory lifecycle management, and on-demand semantic activation.
Table~\ref{tab:jitoma_hyperparameters} summarizes the default values used in
all experiments.

\begin{table}[ht]
    \centering
    \small
    \caption{Default hyperparameters of JITOMA.}
    \label{tab:jitoma_hyperparameters}
    \begin{tabular}{lcc}
        \toprule
        Hyperparameter & Symbol & Value \\
        \midrule
        Observation admission ratio & $r$ & $0.5$ \\
        Heatmap pooling ratio & $\rho$ & $0.2$ \\
        \midrule
        Anchor promotion threshold & $N_{\mathrm{prom}}$ & $8$ observations \\
        Ephemeral expiration threshold & $T_{\mathrm{eph}}$ & $20$ frames \\
        \midrule
        Activation candidates per concept & $K_{\mathrm{act}}$ & $3$ \\
        Heatmap evidence weight & $\lambda$ & $0.10$ \\
        \bottomrule
    \end{tabular}
\end{table}

\paragraph{Task-guided observation admission.}
During active tasks, JITOMA ranks tracked observations according to their
alignment with the task-conditioned heatmap.
Let $H_t^q$ denote the 2D heatmap generated for command $q$ at frame $t$.
For each tracked observation $o$, we compute an admission score

\begin{equation}
    s_t(o,q)
    =
    0.3\,\overline{H}_{t,\mathrm{mask}}^{q}(o)
    +
    0.7\,\overline{H}_{t,\mathrm{top}\text{-}\rho}^{q}(o),
\end{equation}

where the first term represents the average heatmap response inside the
object mask, and the second term captures localized high-response regions
within the top $\rho$ fraction of pixels. We set $\rho=0.2$ in all
experiments.

Given $N_t$ eligible tracked observations at frame $t$, JITOMA retains the
top

\begin{equation}
    K_{o,t}=\lceil rN_t\rceil
\end{equation}

observations according to $s_t(o,q)$, where $K_{o,t}$ denotes the number of
observations admitted into memory and $r=0.5$ is the admission ratio.
This adaptive budget allows the frontend to allocate memory updates according
to the current observation density rather than using a fixed number of
admitted observations.

\paragraph{Task-free conservative writing}
\label{app:conservative_writing}

When no task query is active, JITOMA does not apply task-conditioned
heatmap gating. Instead, it maintains a lightweight task-agnostic writing
policy whose goal is to establish stable object evidence while avoiding
unnecessary repeated memory updates.
Newly observed tracks are admitted when they satisfy the tracking confidence criterion and have not previously contributed observations to memory.
For previously written tracks, JITOMA only performs additional updates when
they provide sufficiently improved visual evidence.
This distinction prevents persistent memory from being dominated by repeated
observations of the same object while still allowing newly emerging or
higher-quality views to refine the stored representation.

For repeated observations, JITOMA evaluates whether a new view provides
meaningful visual improvement using an image-quality proxy.
For an observation $o_{t,i}$, the quality score is defined as

\begin{equation}
Q_{t,i}
=
0.55 A_{t,i}
+
0.15 B_{t,i}
+
0.30 P_{t,i},
\label{eq:observation_visual_quality}
\end{equation}

where $A_{t,i}$, $B_{t,i}$, and $P_{t,i}$ represent normalized object area,
image centeredness, and tracking confidence, respectively.
The score provides a lightweight measure of observation quality without
requiring semantic understanding or additional 3D reconstruction.

Let $e_i(t)$ denote the number of observations previously written for track
$i$, and let $Q_{t-1,i}^{\star}$ denote its best historical quality score.
A previously written track is updated when

\begin{equation}
e_i(t)<3
\lor
\left(
Q_{t-1,i}^{\star}<0.80
\land
Q_{t,i}-Q_{t-1,i}^{\star}\geq0.10
\right).
\label{eq:conservative_repeat_gate}
\end{equation}

The first three writes allow initial evidence accumulation.
After this stage, additional updates are accepted only when the observation
provides a sufficient quality improvement.
To avoid excessive updates across many tracks, at most three repeated tracks
are selected per frame according to update priority.

\paragraph{Memory lifecycle.}
Admitted observations are first maintained as ephemeral hypotheses before
being promoted to persistent dormant anchors.
A hypothesis is promoted when its associated track accumulates
$N_{\mathrm{prom}}=8$ observations, using temporal observation support as a
lightweight persistence criterion rather than performing semantic verification
at this stage.

An unpromoted hypothesis is removed after $T_{\mathrm{eph}}=20$ inactive
frames, where the threshold is measured from the latest observation update.
This expiration rule only applies to ephemeral states; promoted anchors remain
in memory for future task-conditioned activation.
The global memory capacity is fixed to 512 nodes, including both ephemeral
hypotheses and dormant anchors.
Importantly, dormant storage does not imply semantic expansion: captions and
functional structures are generated only after an anchor is activated by a
task query.

\paragraph{On-demand semantic activation.}
When a task query arrives, JITOMA retrieves relevant dormant anchors before
performing expensive semantic expansion.
For each concept $c\in C(q)$, at most $K_{\mathrm{act}}=3$ candidate anchors
are retrieved from $V^{\mathrm{anc}}$ using the CLIP visual key and
task-conditioned evidence:

\begin{equation}
    S(v_j,c)
    =
    \cos\!\left(\operatorname{CLIP}_{\mathrm{text}}(c),z_j\right)
    +
    \lambda \tilde{h}_j^c .
\end{equation}

where $z_j$ denotes the stored visual key of anchor $v_j$ and
$\tilde{h}_j^c$ is the normalized accumulated heatmap evidence for concept
$c$.
We set $\lambda=0.10$, providing a small perceptual correction while keeping
visual-semantic similarity as the dominant retrieval signal.
Only retrieved candidates undergo captioning and subsequent semantic
expansion; all remaining anchors stay dormant.

\subsubsection{Hyperparameter Sensitivity Analysis}
\label{app:hyperparameter_sensitivity}

We investigate the sensitivity of JITOMA to three hyperparameters that
directly control information admission and semantic expansion:
the observation admission ratio $r$, the activation candidate budget
$K_{\mathrm{act}}$, and the heatmap evidence weight $\lambda$.
We perform one-factor-at-a-time experiments on Tier~2 of JITOMA-Bench,
while fixing all remaining parameters to the default configuration
$(r,K_{\mathrm{act}},\lambda)=(0.5,3,0.1)$.
Each configuration is evaluated on the three benchmark scenes with identical
task parser outputs and the same tracking, memory, and geometry processing
settings.

We report mIoU, mR@1, mR@3, and TPF following the main evaluation protocol.
To measure the extent of semantic expansion, we additionally report
Accumulated Objects (Acc.~Obj.), which counts the number of unique object
nodes that receive semantic captions throughout the replay.
Unlike Peak, which measures the maximum instantaneous active semantic-state
size, Acc.~Obj. captures cumulative semantic materialization over time.

\begin{table}[ht]
    \centering
    \small
    \setlength{\tabcolsep}{6pt}
    \caption{
    Hyperparameter sensitivity analysis of JITOMA on Tier~2 of
    JITOMA-Bench.
    Only the parameter indicated in each block is varied while all other
    parameters remain fixed to the default configuration.
    Results are averaged across the three scenes.
    Default settings are marked with $\star$.
    }
    \label{tab:jitoma_hyperparameter_sensitivity}

    \begin{tabular}{@{}lrrrrr@{}}
        \toprule
        Setting
        & mIoU $\uparrow$
        & mR@1 $\uparrow$
        & mR@3 $\uparrow$
        & TPF $\downarrow$
        & Acc.~Obj. $\downarrow$ \\
        \midrule

        \multicolumn{6}{l}{
            \textit{Observation admission ratio}
            ($K_{\mathrm{act}}=3$, $\lambda=0.1$)
        } \\
        $r=0.25$
        & 23.35 & 54.45 & \textbf{67.24} & 0.2781 & 82.67 \\
        $r=0.50^{\star}$
        & \textbf{23.95} & \textbf{56.83} & 67.11 & 0.2794 & 84.33 \\
        $r=0.75$
        & 23.40 & 55.11 & 63.36 & 0.2790 & 84.00 \\
        $r=1.00$
        & 23.18 & 54.76 & 64.59 & 0.2757 & 83.00 \\

        \midrule

        \multicolumn{6}{l}{
            \textit{Activation candidate budget}
            ($r=0.5$, $\lambda=0.1$)
        } \\
        $K_{\mathrm{act}}=1$
        & 22.53 & 51.19 & 51.19 & 0.2624 & 34.67 \\
        $K_{\mathrm{act}}=2$
        & 23.83 & 57.14 & 65.87 & 0.2704 & 63.00 \\
        $K_{\mathrm{act}}=3^{\star}$
        & 23.95 & 56.83 & 67.11 & 0.2794 & 84.33 \\
        $K_{\mathrm{act}}=5$
        & 23.81 & 53.75 & 65.43 & 0.2952 & 109.33 \\
        $K_{\mathrm{act}}=8$
        & \textbf{24.81} & \textbf{58.29} &
          \textbf{68.83} & 0.3124 & 147.00 \\

        \midrule

        \multicolumn{6}{l}{
            \textit{Heatmap evidence weight}
            ($r=0.5$, $K_{\mathrm{act}}=3$)
        } \\
        $\lambda=0$
        & 21.49 & 51.90 & 64.02 & 0.2742 & 58.33 \\
        $\lambda=0.05$
        & 21.09 & 48.32 & 63.10 & 0.2742 & 68.33 \\
        $\lambda=0.10^{\star}$
        & 23.95 & \textbf{56.83} & \textbf{67.11}
        & 0.2794 & 84.33 \\
        $\lambda=0.20$
        & \textbf{24.02} & 56.04 & 66.31
        & 0.2820 & 89.00 \\

        \bottomrule
    \end{tabular}
\end{table}

\paragraph{Observation admission ratio.}
The observation admission ratio controls how much task-conditioned visual
evidence is committed by the frontend.
The default setting $r=0.5$ achieves the strongest overall grounding
performance, obtaining the highest mIoU (23.95) and mR@1 (56.83) among all
tested ratios.
Reducing the ratio to $0.25$ slightly improves mR@3, but decreases mIoU and
mR@1, suggesting that overly aggressive filtering may remove useful
observations.
Increasing the admission ratio beyond $0.5$ does not consistently improve
accuracy, indicating that admitting additional observations alone is
insufficient for better grounding.

The relatively stable Acc.~Obj. and TPF across different values of $r$
suggest that this parameter mainly influences the relevance of admitted
evidence rather than the total semantic expansion cost.
We therefore retain $r=0.5$ as a balanced operating point between preserving
task-related observations and limiting unnecessary memory updates.

\paragraph{Activation candidate budget.}
The activation candidate budget $K_{\mathrm{act}}$ controls the amount of
semantic expansion performed after task activation and introduces the clearest
accuracy--efficiency trade-off.
A larger candidate budget provides more opportunities to recover relevant
anchors, but also increases the number of objects exposed to semantic
processing.

Increasing $K_{\mathrm{act}}$ from 3 to 8 improves mR@1 and mR@3 from
56.83/67.11 to 58.29/68.83.
However, this improvement increases Acc.~Obj. from 84.33 to 147.00 and raises
TPF from 0.2794s to 0.3124s.
Conversely, smaller budgets reduce semantic expansion but provide fewer
candidate anchors for retrieval.
Although $K_{\mathrm{act}}=8$ achieves slightly higher accuracy, we select
$K_{\mathrm{act}}=3$ as the default configuration because it provides strong
grounding performance while maintaining bounded on-demand semantic growth.

\paragraph{Heatmap evidence weight.}
The heatmap evidence weight $\lambda$ determines the contribution of
task-conditioned perceptual evidence during anchor ranking.
Removing this term ($\lambda=0$) decreases mIoU, mR@1, and mR@3 by
2.46, 4.94, and 3.09 percentage points compared with the default setting,
respectively.
This demonstrates that task-conditioned evidence complements CLIP-based
visual similarity during anchor activation.

Increasing $\lambda$ to 0.2 provides only marginal accuracy improvement while
slightly increasing semantic expansion.
The results indicate that task evidence is most effective as a complementary
signal rather than a replacement for visual-semantic matching.
We therefore retain $\lambda=0.1$ as a moderate weighting factor.

Overall, these sensitivity experiments support the default configuration
$(r,K_{\mathrm{act}},\lambda)=(0.5,3,0.1)$ as a balanced operating point,
providing strong grounding accuracy while preserving the bounded semantic
growth behavior targeted by JITOMA.

%% file: iclr2027_conference.bib
@inproceedings{gu2024conceptgraphs,
  title={Conceptgraphs: Open-vocabulary 3d scene graphs for perception and planning},
  author={Gu, Qiao and Kuwajerwala, Ali and Morin, Sacha and Jatavallabhula, Krishna Murthy and Sen, Bipasha and Agarwal, Aditya and Rivera, Corban and Paul, William and Ellis, Kirsty and Chellappa, Rama and others},
  booktitle={2024 IEEE International Conference on Robotics and Automation (ICRA)},
  pages={5021--5028},
  year={2024},
  organization={IEEE}
}

@inproceedings{werby2024hierarchical,
  title={Hierarchical open-vocabulary 3d scene graphs for language-grounded robot navigation},
  author={Werby, Abdelrhman and Huang, Chenguang and B{\"u}chner, Martin and Valada, Abhinav and Burgard, Wolfram},
  booktitle={First Workshop on Vision-Language Models for Navigation and Manipulation at ICRA 2024},
  year={2024}
}

@inproceedings{koch2024open3dsg,
  title={Open3dsg: Open-vocabulary 3d scene graphs from point clouds with queryable objects and open-set relationships},
  author={Koch, Sebastian and Vaskevicius, Narunas and Colosi, Mirco and Hermosilla, Pedro and Ropinski, Timo},
  booktitle={Proceedings of the IEEE/CVF Conference on Computer Vision and Pattern Recognition},
  pages={14183--14193},
  year={2024}
}

@article{jatavallabhula2023conceptfusion,
  title={Conceptfusion: Open-set multimodal 3d mapping},
  author={Jatavallabhula, Krishna Murthy and Kuwajerwala, Alihusein and Gu, Qiao and Omama, Mohd and Chen, Tao and Maalouf, Alaa and Li, Shuang and Iyer, Ganesh and Saryazdi, Soroush and Keetha, Nikhil and others},
  journal={arXiv preprint arXiv:2302.07241},
  year={2023}
}

@inproceedings{linok2025beyond,
  title={Beyond bare queries: Open-vocabulary object grounding with 3d scene graph},
  author={Linok, Sergey and Zemskova, Tatiana and Ladanova, Svetlana and Titkov, Roman and Yudin, Dmitry and Monastyrny, Maxim and Valenkov, Aleksei},
  booktitle={2025 IEEE International Conference on Robotics and Automation (ICRA)},
  pages={13582--13589},
  year={2025},
  organization={IEEE}
}

@article{maggio2024clio,
  title={Clio: Real-time task-driven open-set 3d scene graphs},
  author={Maggio, Dominic and Chang, Yun and Hughes, Nathan and Trang, Matthew and Griffith, Dan and Dougherty, Carlyn and Cristofalo, Eric and Schmid, Lukas and Carlone, Luca},
  journal={IEEE Robotics and Automation Letters},
  year={2024},
  publisher={IEEE}
}

@inproceedings{yamazaki2024open,
  title={Open-fusion: Real-time open-vocabulary 3d mapping and queryable scene representation},
  author={Yamazaki, Kashu and Hanyu, Taisei and Vo, Khoa and Pham, Thang and Tran, Minh and Doretto, Gianfranco and Nguyen, Anh and Le, Ngan},
  booktitle={2024 IEEE International Conference on Robotics and Automation (ICRA)},
  pages={9411--9417},
  year={2024},
  organization={IEEE}
}

@article{yan2025dynamic,
  title={Dynamic open-vocabulary 3d scene graphs for long-term language-guided mobile manipulation},
  author={Yan, Zhijie and Li, Shufei and Wang, Zuoxu and Wu, Lixiu and Wang, Han and Zhu, Jun and Chen, Lijiang and Liu, Jihong},
  journal={IEEE Robotics and Automation Letters},
  year={2025},
  publisher={IEEE}
}

@inproceedings{fungraph3d,
  title={Open-vocabulary functional 3d scene graphs for real-world indoor spaces},
  author={Zhang, Chenyangguang and Delitzas, Alexandros and Wang, Fangjinhua and Zhang, Ruida and Ji, Xiangyang and Pollefeys, Marc and Engelmann, Francis},
  booktitle={Proceedings of the Computer Vision and Pattern Recognition Conference},
  pages={19401--19413},
  year={2025}
}

@inproceedings{chang2026rag,
  title={Rag-3dsg: Enhancing 3d scene graphs with re-shot guided retrieval-augmented generation},
  author={Chang, Yue and Chen, Rufeng and Zhang, Zhaofan and Chen, Yi and Tian, Yifan and Xie, Sihong},
  booktitle={European Conference on Computer Vision},
  pages={223--240},
  year={2026},
  organization={Springer}
}

@article{ju2025momagraph,
  title={MomaGraph: State-Aware Unified Scene Graphs with Vision-Language Model for Embodied Task Planning},
  author={Ju, Yuanchen and Liang, Yongyuan and Wang, Yen-Jen and Gireesh, Nandiraju and Ju, Yuanliang and Lee, Seungjae and Gu, Qiao and Hsieh, Elvis and Huang, Furong and Sreenath, Koushil},
  journal={arXiv preprint arXiv:2512.16909},
  year={2025}
}

@article{buchner2026articulated,
  title={Articulated 3D Scene Graphs for Open-World Mobile Manipulation},
  author={B{\"u}chner, Martin and R{\"o}fer, Adrian and Engelbracht, Tim and Welschehold, Tim and Bauer, Zuria and Blum, Hermann and Pollefeys, Marc and Valada, Abhinav},
  journal={arXiv preprint arXiv:2602.16356},
  year={2026}
}

@inproceedings{rotondi2025fungraph,
  title={Fungraph: Functionality aware 3d scene graphs for language-prompted scene interaction},
  author={Rotondi, Dennis and Scaparro, Fabio and Blum, Hermann and Arras, Kai O},
  booktitle={2025 IEEE/RSJ International Conference on Intelligent Robots and Systems (IROS)},
  pages={4083--4090},
  year={2025},
  organization={IEEE}
}

@article{hughes2022hydra,
  title={Hydra: A real-time spatial perception system for 3D scene graph construction and optimization},
  author={Hughes, Nathan and Chang, Yun and Carlone, Luca},
  journal={arXiv preprint arXiv:2201.13360},
  year={2022}
}

@article{rosinol2021kimera,
  title={Kimera: From SLAM to spatial perception with 3D dynamic scene graphs},
  author={Rosinol, Antoni and Violette, Andrew and Abate, Marcus and Hughes, Nathan and Chang, Yun and Shi, Jingnan and Gupta, Arjun and Carlone, Luca},
  journal={The International Journal of Robotics Research},
  volume={40},
  number={12-14},
  pages={1510--1546},
  year={2021},
  publisher={SAGE Publications Sage UK: London, England}
}

@inproceedings{wu2021scenegraphfusion,
  title={Scenegraphfusion: Incremental 3d scene graph prediction from rgb-d sequences},
  author={Wu, Shun-Cheng and Wald, Johanna and Tateno, Keisuke and Navab, Nassir and Tombari, Federico},
  booktitle={Proceedings of the IEEE/CVF Conference on Computer Vision and Pattern Recognition},
  pages={7515--7525},
  year={2021}
}

@article{maggio2026found,
  title={FOUND-IT: Foundation-model-first Task-driven 3D Scene Graphs with Granularity on Demand},
  author={Maggio, Dominic and Gorlo, Nicolas and Carlone, Luca},
  journal={arXiv preprint arXiv:2605.25371},
  year={2026}
}

@inproceedings{agia2022taskography,
  title={Taskography: Evaluating robot task planning over large 3d scene graphs},
  author={Agia, Christopher and Jatavallabhula, Krishna Murthy and Khodeir, Mohamed and Miksik, Ondrej and Vineet, Vibhav and Mukadam, Mustafa and Paull, Liam and Shkurti, Florian},
  booktitle={Conference on Robot Learning},
  pages={46--58},
  year={2022},
  organization={PMLR}
}

@article{maggio2025bayesian,
  title={Bayesian fields: Task-driven open-set semantic Gaussian splatting},
  author={Maggio, Dominic and Carlone, Luca},
  journal={arXiv preprint arXiv:2503.05949},
  year={2025}
}

@inproceedings{gorlo2026describe,
  title={Describe anything anywhere at any moment},
  author={Gorlo, Nicolas and Schmid, Lukas and Carlone, Luca},
  booktitle={Proceedings of the IEEE/CVF Conference on Computer Vision and Pattern Recognition},
  pages={35002--35013},
  year={2026}
}

@article{gadre2022clip,
  title={Clip on wheels: Zero-shot object navigation as object localization and exploration},
  author={Gadre, Samir Yitzhak and Wortsman, Mitchell and Ilharco, Gabriel and Schmidt, Ludwig and Song, Shuran},
  journal={arXiv preprint arXiv:2203.10421},
  volume={3},
  number={4},
  pages={7},
  year={2022}
}

@inproceedings{shah2023lm,
  title={Lm-nav: Robotic navigation with large pre-trained models of language, vision, and action},
  author={Shah, Dhruv and Osi{\'n}ski, B{\l}a{\.z}ej and Levine, Sergey and others},
  booktitle={Conference on robot learning},
  pages={492--504},
  year={2023},
  organization={PMLR}
}

@article{yin2024sg,
  title={Sg-nav: Online 3d scene graph prompting for llm-based zero-shot object navigation},
  author={Yin, Hang and Xu, Xiuwei and Wu, Zhenyu and Zhou, Jie and Lu, Jiwen},
  journal={Advances in neural information processing systems},
  volume={37},
  pages={5285--5307},
  year={2024}
}

@article{chen2026psg,
  title={PSG-Nav: Probabilistic Scene Graph Navigation via Multiverse Decision Making},
  author={Chen, Rufeng and Chang, Yue and Tang, Xiaqiang and Chen, Hechang and Xie, Sihong},
  journal={arXiv preprint arXiv:2606.01313},
  year={2026}
}

@article{xia2026exploring,
  title={Exploring Bottlenecks in VLM-LLM Navigation: How 3D Scene Understanding Capability Impacts Zero-Shot VLN},
  author={Xia, Ziyi and Xiong, Chaoran and Wei, Litao and Hu, Xinhao and Pei, Ling},
  journal={arXiv preprint arXiv:2605.14801},
  year={2026}
}

@inproceedings{shridhar2022cliport,
  title={Cliport: What and where pathways for robotic manipulation},
  author={Shridhar, Mohit and Manuelli, Lucas and Fox, Dieter},
  booktitle={Conference on robot learning},
  pages={894--906},
  year={2022},
  organization={PMLR}
}

@inproceedings{rashid2023language,
  title={Language embedded radiance fields for zero-shot task-oriented grasping},
  author={Rashid, Adam and Sharma, Satvik and Kim, Chung Min and Kerr, Justin and Chen, Lawrence Yunliang and Kanazawa, Angjoo and Goldberg, Ken},
  booktitle={7th Annual Conference on Robot Learning},
  year={2023}
}

@article{honerkamp2024language,
  title={Language-grounded dynamic scene graphs for interactive object search with mobile manipulation},
  author={Honerkamp, Daniel and B{\"u}chner, Martin and Despinoy, Fabien and Welschehold, Tim and Valada, Abhinav},
  journal={IEEE Robotics and Automation Letters},
  year={2024},
  publisher={IEEE}
}

@article{puigjaner2026relationship,
  title={Relationship-Aware Hierarchical 3D Scene Graph for Task Reasoning},
  author={Puigjaner, Albert Gassol and Zacharia, Angelos and Alexis, Kostas},
  journal={arXiv preprint arXiv:2602.02456},
  year={2026}
}

@misc{qwen3.5,
    title  = {{Qwen3.5}: Towards Native Multimodal Agents},
    author = {{Qwen Team}},
    month  = {February},
    year   = {2026},
    url    = {https://qwen.ai/blog?id=qwen3.5}
}

@inproceedings{clipseg,
  title={Image segmentation using text and image prompts},
  author={L{\"u}ddecke, Timo and Ecker, Alexander},
  booktitle={Proceedings of the IEEE/CVF conference on computer vision and pattern recognition},
  pages={7086--7096},
  year={2022}
}

@inproceedings{dam,
  title={Describe anything: Detailed localized image and video captioning},
  author={Lian, Long and Ding, Yifan and Ge, Yunhao and Liu, Sifei and Mao, Hanzi and Li, Boyi and Pavone, Marco and Liu, Ming-Yu and Darrell, Trevor and Yala, Adam and others},
  booktitle={Proceedings of the IEEE/CVF International Conference on Computer Vision},
  pages={21766--21777},
  year={2025}
}

@misc{openai2026gpt54,
  author       = {{OpenAI}},
  title        = {Introducing {GPT-5.4}},
  year         = {2026},
  month        = mar,
  howpublished = {\url{https://openai.com/index/introducing-gpt-5-4/}},
  note         = {Accessed: July 14, 2026}
}
